\documentclass{article}
\usepackage{arxiv}
\usepackage[utf8]{inputenc}
\usepackage[T1]{fontenc}
\usepackage{hyperref}
\usepackage{url}
\usepackage{booktabs}
\usepackage{amsfonts}
\usepackage{nicefrac}
\usepackage{microtype}
\usepackage{graphicx}
\graphicspath{ {./images/} }
\usepackage{amsmath, amssymb}
\usepackage{lineno}
\usepackage[numbers]{natbib}
\usepackage{tikz}
\usetikzlibrary{arrows.meta,positioning,shapes.geometric}
\usepackage{calc}
\usepackage{xcolor}
\usepackage{adjustbox}
\usepackage{todonotes}
\usepackage{subcaption}
\usepackage{pifont}
\usepackage{threeparttable}
\usepackage{multirow}
\usepackage{makecell}
\usepackage{siunitx}
\usepackage{caption}
\usepackage{nameref}
\usepackage[fixed]{fontawesome5}
\usepackage[misc]{ifsym}

\DeclareMathOperator*{\argmin}{argmin}
\DeclareMathOperator*{\argmax}{argmax}

\title{Deep Generative Classification of Blood Cell Morphology}

\author{
  Simon Deltadahl$^1$ \And
  Julian Gilbey$^1$ \And
  Christine Van Laer$^{2,3}$ \And
  Nancy Boeckx$^{2,4}$ \And
  Mathie Leers$^5$ \And
  Tanya Freeman$^6$ \And
  Laura Aiken$^6$ \And
  Timothy Farren$^6$ \And
  Matthew Smith$^6$ \And
  Mohamad Zeina$^{7}$ \And
  BloodCounts! consortium \And
  James HF Rudd$^{10}$ \And
  Concetta Piazzese$^{6,8}$ \And
  Joseph Taylor$^{6,8}$ \And
  Nicholas Gleadall$^9$ \And
  Carola-Bibiane Schönlieb$^1$ \And
  Suthesh Sivapalaratnam$^{6,8,\dag}$ \And
  Michael Roberts$^{1,10,\faEnvelope,\dag}$ \And
  Parashkev Nachev$^{7,\dag}$
}

\begin{document}
\maketitle

\begin{abstract}
Accurate classification of haematological cells is critical for diagnosing blood disorders, but presents significant challenges for machine automation owing to the complexity of cell morphology, heterogeneities of biological, pathological, and imaging characteristics, and the imbalance of cell type frequencies. We introduce CytoDiffusion, a diffusion-based classifier that effectively models blood cell morphology, combining accurate classification with robust anomaly detection, resistance to distributional shifts, interpretability, data efficiency, and superhuman uncertainty quantification. Our approach outperforms state-of-the-art discriminative models in anomaly detection (AUC 0.990 vs.\ 0.918), resistance to domain shifts (85.85\% vs.\ 74.38\% balanced accuracy), and performance in low-data regimes (95.88\% vs.\ 94.95\% balanced accuracy). Notably, our model generates synthetic blood cell images that are nearly indistinguishable from real images, as demonstrated by an authenticity test in which expert haematologists achieved only 52.3\% accuracy (95\% CI: [50.5\%, 54.2\%]) in distinguishing real from generated images. Furthermore, we enhance model explainability through the generation of directly interpretable counterfactual heatmaps. Our comprehensive evaluation framework, encompassing these multiple performance dimensions, establishes a new benchmark for medical image analysis in haematology, ultimately enabling improved diagnostic accuracy in clinical settings. Our code is available at \url{https://github.com/CambridgeCIA/CytoDiffusion}.
\end{abstract}

\vfill
$^1$Department of Applied Mathematics and Theoretical Physics, University of Cambridge, Cambridge, UK\\
$^2$Department of Laboratory Medicine, UZ Leuven, Leuven, Belgium\\
$^3$Department of Cardiovascular Sciences, Center for Molecular and Vascular Biology, University of Leuven, Leuven, Belgium\\
$^4$Department of Oncology, KU Leuven, Leuven, Belgium\\
$^5$Zuyderland Medical Center, Sittard-Geleen, Netherlands\\
$^6$Barts Health NHS Trust, London, United Kingdom\\
$^{7}$Queen Square Institute of Neurology, University College London, London, UK\\
$^8$Queen Mary University of London, London, UK\\
$^9$Department of Haematology, University of Cambridge, Cambridge, UK\\
$^{10}$Department of Medicine, University of Cambridge, Cambridge, UK\\
$^{\faEnvelope}$Corresponding author: mr808@cam.ac.uk\\
$^\dag$Equal contribution\\

\maketitle
\clearpage

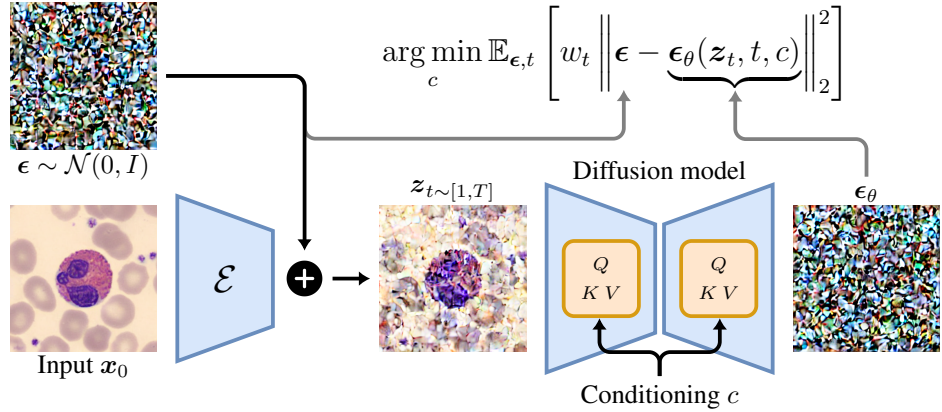
\begin{figure}[htb]
    \centering
    \begin{adjustbox}{center}
    \begin{tikzpicture}
    \definecolor{light_blue}{rgb}{0.847, 0.906, 0.984}
    \definecolor{blue}{rgb}{0.412, 0.553, 0.741}

    \node (noise1) {\includegraphics[width=0.118\textwidth]{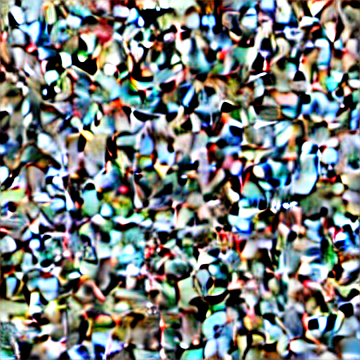}};

    \node[below=of noise1, yshift=0.5cm] (original) {\includegraphics[width=0.118\textwidth]{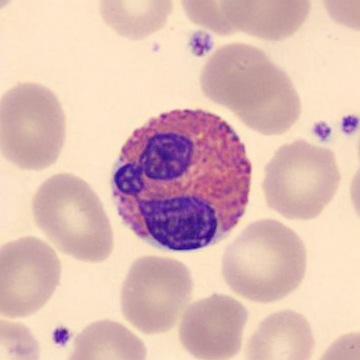}};

    \node[below=of noise1, yshift=1.2cm] (label_noise1) {$\boldsymbol{\epsilon} \sim \mathcal{N}(0,I)$};

    \node[below=of original, yshift=1.2cm] (label_original) {Input $\boldsymbol{x}_0$};

    \node[right=of original, xshift=-1.0cm, yshift=0.0cm] (VAE_encoder) {
    \begin{tikzpicture}
    \node[trapezium, trapezium angle=70, minimum width=2.2cm, minimum height=0.47cm, draw=blue, fill=light_blue, line width=0.5mm, shape border rotate=270] (encoder_SD) {};
    \end{tikzpicture}
    };

    \node[at=(VAE_encoder.center)] (VAE_encoder_label) {\Large $\mathcal{E}$};

    \node[right=of VAE_encoder, xshift=-1.1cm, yshift=0.0cm] (plus) {\includegraphics[width=0.03\textwidth]{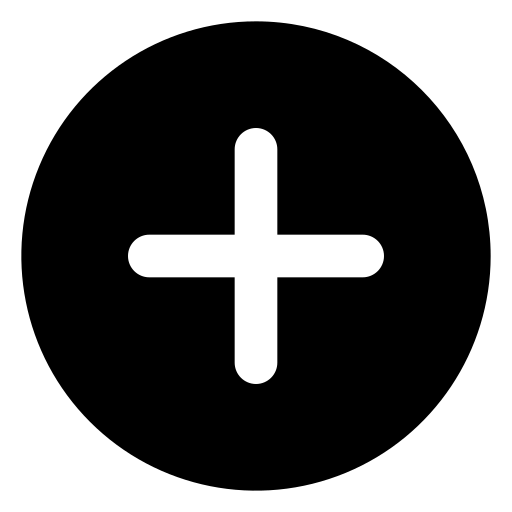}};

    \node[right=of plus, xshift=-0.50cm] (combined) {\includegraphics[width=0.118\textwidth]{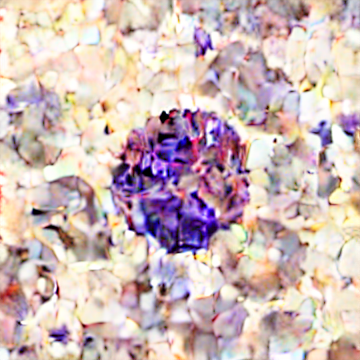}};

    \node[above=of combined, yshift=-1.2cm] (xt) {$\boldsymbol{z}_{t\sim [1,T]}$};

    \node[right=of combined, xshift=-1.0cm, yshift=0.0cm] (stable_diffusion) {
    \begin{tikzpicture}
        \definecolor{boxfill}{RGB}{255, 230, 205}
        \definecolor{boxedge}{RGB}{216, 155, 18}

        \node[trapezium, trapezium angle=70, minimum width=2.5cm, minimum height=0.47cm, draw=blue, fill=light_blue, line width=0.5mm, shape border rotate=270] (encoder_SD) {};
        \node[trapezium, trapezium angle=70, minimum width=2.5cm, minimum height=0.47cm, draw=blue, fill=light_blue, line width=0.5mm, shape border rotate=90, right=0.067cm of encoder_SD] (decoder_SD) {};

        \node[rectangle, rounded corners, draw=boxedge, fill=boxfill, minimum width=1.0cm, minimum height=1.0cm, align=center, line width=0.5mm, at=(encoder_SD.center), xshift=-0.52cm] (QKV_e) {\scriptsize $Q$\\\scriptsize$K$ $V$};
        \node[rectangle, rounded corners, draw=boxedge, fill=boxfill, minimum width=1.0cm, minimum height=1.0cm, align=center, line width=0.5mm, at=(decoder_SD.center), xshift=-0.52cm] (QKV_d) {\scriptsize $Q$\\\scriptsize$K$ $V$};

    \end{tikzpicture}
    };
    \node[below=of stable_diffusion, yshift=1.1cm, xshift=0.0cm] (class_conditioning) {Conditioning $c$};

    \draw[-{Latex[length=1.5mm, width=2mm]}, line width=0.5mm, rounded corners=4pt]
        (class_conditioning)
        -- ++(0,0.6)
        -- ++(-0.8,0)
        -- ++(0,0.45);
    \draw[-{Latex[length=1.5mm, width=2mm]}, line width=0.5mm, rounded corners=4pt]
        (class_conditioning)
        -- ++(0,0.6)
        -- ++(0.8,0)
        -- ++(0,0.45);

    \node[above=of stable_diffusion, yshift=-1.2cm] (SD_label) {Diffusion model};

    \node[right=of noise1, yshift=0.3cm, xshift=1.75cm] (equation) {\large $\underset{c}{\mathrm{argmin}} \, \mathbb{E}_{\boldsymbol{\epsilon}, t} \left[ w_t \left\| \boldsymbol{\epsilon} - \underbrace{\boldsymbol{\epsilon}_\theta (\boldsymbol{z}_t, t, c)} \right\|^2_2 \right] $};

    \node[right=of stable_diffusion, xshift=-1.0cm] (noise2) {\includegraphics[width=0.118\textwidth]{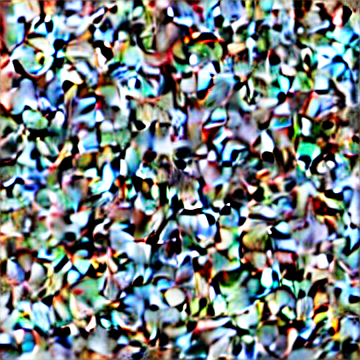}};

    \node[above=of noise2, yshift=-1.2cm] (label_noise2) {$\boldsymbol{\epsilon}_\theta$};

    \draw[-{Latex[length=1.5mm, width=2mm]}, line width=0.5mm, rounded corners=4pt, draw=gray] (noise1.east) ++(0,-0.0) -- ++(1.845,0) -- ++(0,-0.835) -- ++(4.16,0) -- ++(0,0.6);

    \draw[-{Latex[length=1.5mm, width=2mm]}, line width=0.5mm, rounded corners=4pt] (noise1.east) ++(0,0.0) -| (plus);
    \draw[-{Latex[length=1.5mm, width=2mm]}, line width=0.5mm, rounded corners=4pt] (plus) -- (combined);

    \draw[-{Latex[length=1.5mm, width=2mm]}, line width=0.5mm, rounded corners=4pt, draw=gray] (label_noise2.north) -- ++(0,0.5) -- ++(-1.83,0) -- ++(0,0.6);

    \end{tikzpicture}
    \end{adjustbox}
    \caption{Representation of the diffusion-based classification process. An input image $\boldsymbol{x}_0$ is first encoded into a latent space using an encoder $\mathcal{E}$. Gaussian noise $\boldsymbol{\epsilon} \sim \mathcal{N}(0,\boldsymbol{I})$ is then added to create a noisy latent representation $\boldsymbol{z}_t$. This noisy representation is fed through a diffusion model for each possible class condition $c$. The model predicts the noise $\boldsymbol{\epsilon}_\theta$ for each condition. The classification decision is made by selecting the class that minimises the error between the predicted noise $\boldsymbol{\epsilon}_\theta$ and the true noise $\boldsymbol{\epsilon}$.}
    \label{fig:diffusion_classifier}
\end{figure}

The haematological is amongst the most complex of physiological systems, and uniquely intertwined with all others. Though often quantified by simple ``blood counts'' of cell class frequencies, its characteristics are both supremely rich and highly variable within and across individuals~\cite{bain2021blood}.
Characterising the morphological appearances of individual blood cells as seen on light microscopy is often critical to managing haematological disorders. Complex modulation of cell morphology by diverse biological, pathological, and instrumental factors, requires that this task necessarily be performed by trained experts. Moreover, labelling cells by their major morphological type, e.g. lymphocyte, is only the crudest form of description, on which finer fractionation into subtypes, across a wide spectrum of (ab)normality, is overlaid.

Indeed, the task of morphological characterisation is both open-ended and lacks a definitive ground truth: there may be morphological patterns whose subtlety has concealed great clinical significance, and some morphological classes are purely expert-determined visual phenotypes with no means of objective corroboration. Moreover, pathological appearances may be highly unusual or unique, precluding classification into any class, even at the simplest level of description, and such anomalies should be explicitly identified.
The difficulty is commonly compounded by interactions with irrelevant biological features with variable representation across the population, and instrumental variations of technical origin~\cite{Kratz_2019, Buttarello_2008}. The challenge, in short, is one human experts can only imperfectly meet, inevitably exhibiting marked variation with skill and experience~\cite{van_de_Geijn_van_2011, Metter_Nathwani_1985} and therefore training of ML-based models to automate morphological characterisation is intrinsically difficult.

So framed, the task of automating blood morphological analysis has a different aim from that commonly assumed, needs broader validation, and is harder to accomplish. The primary aim is not to approximate a human expert in a more cost-effective, reproducible, and scalable alternative, but to capture the space of possible morphological appearances with potentially superhuman fidelity, flexibility, and metacognitive awareness. It cannot plausibly be achieved by a discriminative model trained to classify cells into standard morphological classes, for such an approach can only approximate human fidelity, and is ill-equipped to deal with the domain shifts, class imbalances and anomalies, complex interactions between biological, pathological, and instrumental features, and metacognitive demands, that the foregoing implies. Nor can success in the task be evaluated by standard measures of in-distribution performance, for they are remote from real-world experience.

A desirable automated cell characterisation model would have the following five key properties. Firstly, it should be robust to domain shift and generalise to different biological, pathological, and instrumental contexts and class distributions~\cite{Yoon_Oh_2024, Koh_Sagawa_2021}. Secondly, the model should achieve high data efficiency, performing well despite sparse ground truth labels and restricted access to comprehensive datasets as commonly found in clinical applications. Thirdly, the decision of the model should aim to be interpretable where possible, as the reasoning behind a model's decisions may be as important as the decisions themselves~\cite{Holzinger_Langs_2019, Rudin_2019}. Fourth, the model should have the ability to identify rare or previously unseen patterns of features as such cases fall outside the model's competence and must be so highlighted. This is particularly important in clinical applications yet often overlooked in model development and evaluation~\cite{Kazerouni_Aghdam_2023}. Finally, the model should be able to quantify the uncertainty attached to its decision, a feature often neglected in model assessments~\cite{Begoli_Bhattacharya_Kusnezov_2019}.

Optimally performing discriminative ML classification models are able to approximate human performance at classifying cells into a priori defined morphological classes. However, they are not inherently designed with desirable properties such as tolerating domain shifts, class imbalances, anomalies and complex interactions between the outcomes and biological, pathological, and instrumental features.
Numerous studies have reported high accuracy in various medical image analysis tasks~\cite{Asghar_Kumar_Hynds_Shaukat, kumar2024medical}, but we show that none exhibit the desirable properties we described and these open challenges remain to be addressed, limiting the clinical applicability of such models.
Therefore, we introduce CytoDiffusion, a modelling approach whose primary aim is not simply to approximate a human expert in a more cost-effective, reproducible, and scalable way, but to capture the space of possible morphological appearances with high-fidelity using a diffusion-based generative model paired with a novel classification approach.

CytoDiffusion is developed and applied to real-world clinical challenges, building on recent work~\cite{Generative_Classifiers, Chen_Dong_Wang_Yang_Duan_Su_Zhu_2024, Clark_Jaini_2023, Li_Prabhudesai_Duggal_Brown_Pathak_2023} using generative models for classification challenges.
A key advantage of generative classifiers is their ability to model the full data distribution rather than merely a decision boundary~\cite{Generative_Classifiers}
and these have shown promising robustness to distribution shifts and ability to avoid shortcut solutions.
In the context of blood cell image classification, CytoDiffusion is compelled to learn the complete morphological characteristics of each cell type, rather than focusing on easily identified but potentially unreliable features. Figure \ref{fig:diffusion_classifier} illustrates the proposed modelling approach and the contributions of this work are: (a) a novel application of latent diffusion models for blood cell image classification, (b) an evaluation framework that goes beyond accuracy and other standard metrics, incorporating domain shift robustness, anomaly detection capability, and performance in low-data regimes, (c) a new dataset of blood cell images that includes artefacts and labeller confidence scores, addressing key limitations in existing datasets, (d) a principled framework for the evaluation of model and human confidence based on established psychometric modelling techniques and (e) a method for generating interpretable heatmaps to explain model's decisions.

Through these contributions, we aim to establish a new standard for the development and assessment of blood cell image classification models. Our work addresses several important aspects of clinical applicability, including robustness, interpretability, and reliability.
We propose that the research community adopt these evaluation tasks and metrics when assessing new models for blood cell image classification. By going beyond simple discriminative statistics, and other conventional benchmarks, we can develop models that are not only high-performing but also trustworthy and clinically relevant.

\section*{Results}
We begin by validating the quality of images generated by CytoDiffusion through an authenticity test. Next, we assess the model's in-domain performance on standard classification tasks across multiple datasets. We then examine CytoDiffusion's ability to quantify uncertainty, comparing its metacognitive capabilities with those of human experts. Following this, we evaluate the model's proficiency in anomaly detection, crucial for identifying rare or unseen cell types. We proceed to test the model's robustness to domain shifts, simulating real-world variability in imaging conditions. Subsequently, we investigate CytoDiffusion's efficiency in low-data regimes, a critical consideration for medical applications where large, well-annotated datasets may be scarce. Finally, we demonstrate CytoDiffusion's explainability through the generation of counterfactual heatmaps, providing interpretable insights into its decision-making process.

\subsection*{CytoDiffusion Generates Images Indistinguishable From Real Images}
To assess whether our model could effectively capture the distribution of blood cell images, we designed an authenticity test.
CytoDiffusion was trained on a dataset comprising 32,619 images. The results of the test, which involved expert haematologists assessing the authenticity of the images, underscore the model's capability to replicate the complex morphological characteristics of blood cells.
The ten haematologists participating in the study evaluated 288 images each, resulting in 2,880 individual assessments. They demonstrated an overall accuracy of 52.3\%, 95\% CI: [50.5\%, 54.2\%] in distinguishing between real and synthetic images, with a sensitivity of 55.8\% and a specificity of 48.9\%. This performance is comparable to random guessing, indicating that the synthetic images produced by CytoDiffusion are almost indistinguishable from real blood cell images, even to experienced professionals.

In addition to the authenticity test, we evaluated the quality of conditional synthesis for each blood cell type. The haematologists' classification of the synthetic images aligned with the model's intended cell type with an agreement rate of 98.6\%. Overall, these results affirm that CytoDiffusion not only generates high-quality synthetic images but also accurately captures the class-defining morphological features of blood cells (see Supplementary Figure~\ref{fig:synthetic-blood-cells}).

\subsection*{CytoDiffusion Demonstrates Competitive Classification Performance}
To establish a baseline for our diffusion-based classifier, we evaluated CytoDiffusion's performance on standard in-domain classification tasks. We used three datasets for this evaluation: CytoData (our custom dataset), Raabin-WBC~\cite{Kouzehkanan_Saghari__2022}, and PBC~\cite{Acevedo} (see Methods for details). Table~\ref{tab:comparison} presents a comparison of the performance of our model against other competitor methods for these datasets.
CytoDiffusion demonstrates competitive performance across all datasets, particularly in balanced accuracy. This metric is especially relevant given the significant class imbalances present in these datasets. For instance, CytoData's class distribution ranges from 103 to 1425 images per class, while Raabin-WBC's test set contains between 89 to 2660 images per class.

\begin{table}[ht]
\centering
\small
\begin{tabular}{clccc}
\toprule
\textbf{Dataset} & \textbf{Method} & \textbf{Year} & \textbf{Accuracy (\%)} & \makecell{\textbf{Balanced} \\ \textbf{Accuracy (\%)}} \\
\midrule
\multirow{3}{*}{\rotatebox[origin=c]{0}{CytoData}}
 & ViT-B/16 & 2024 & 84.42 & 76.89 \\
 & EfficientNetV2-M & 2024 & \underline{85.74} & \underline{80.69} \\
 & CytoDiffusion (ours) & 2024 & \textbf{86.48} & \textbf{85.27} \\
\midrule
\multirow{9}{*}{\rotatebox[origin=c]{0}{\parbox[c]{2cm}{\centering Raabin-WBC\ Test-A}}}
 & Tavakoli et al. \cite{Tavakoli_Ghaffari_2021} & 2021 & 94.65 & 92.76 \\
 & Kouzehkanan et al. \cite{Kouzehkanan_Saghari__2022} & 2022 & \textbf{99.17} & 97.95 \\
 & Chen et al. \cite{Chen_Liu_Hua_2022} & 2022 & 98.71 & \textbf{98.42} \\
 & Jiang et al. \cite{L_C_H_2022} & 2022 & 95.17 & 91.71 \\
 & Rivas et al. \cite{Rivas-Posada_2023} & 2023 & 98.29 & 97.69 \\
 & Tsutsui et al. \cite{Tsutsui_Su_Wen_2023} & 2023 & 98.75 & -\\
 & Li et al. \cite{Li_Liu_2024} & 2024 & 97.80 & -\\
 & ViT-B/16 & 2024 & 96.93 & 97.28 \\
 & EfficientNetV2-M & 2024 & \underline{98.99} & \underline{98.40} \\
 & CytoDiffusion (ours) & 2024 & 96.43 & 97.93 \\
\midrule
\multirow{10}{*}{\rotatebox[origin=c]{0}{PBC}}
 & Acevedo et al. \cite{Acevedo} & 2019 & 96.20 & 96.16 \\
 & Ucar \cite{Ucar_2020} & 2020 & 97.94 & 97.94 \\

 & Rastogi et al. \cite{Rastogi_Khanna_2022} & 2022 & 92.00 & 91.46 \\
 & Ali et al. \cite{Abou_Ali_2023} & 2023 & 93.00 & 91.00 \\
 & Manzari et al. \cite{Manzari_2023} & 2023 & 95.40 & - \\
 & Almalik et al. \cite{Almalik_Alkhunaizi_2023} & 2023 & 93.60 & -\\
 & Tummala et al. \cite{Tummala_Suresh_2023} & 2023 & 97.25 & 97.25 \\
 & Fırat et al. \cite{Firat_2024} & 2024 & \textbf{99.72} & 98.91\\
 & ViT-B/16 & 2024 & 98.70 & 98.76 \\
 & EfficientNetV2-M & 2024 & 99.17 & \underline{99.14} \\
 & CytoDiffusion (ours) & 2024 & \underline{99.35} & \textbf{99.37} \\
\bottomrule
\end{tabular}
\vspace{0.1in}
\caption{Comparison of competitor methods on the CytoData, Raabin-WBC (Test-A) and the PBC dataset. Best results are in bold, second-best are underlined. Balanced accuracy is reported where available.}
\label{tab:comparison}
\end{table}

\subsection*{CytoDiffusion Exhibits Superhuman Uncertainty Quantification} \label{Results Uncertainty Measure}

The biological realm is characterised by constitutional, incompletely reducible uncertainty. In every task, it is valuable to quantify not only the fidelity but also the uncertainty of the agent: human or machine. Metacognitive measures of this kind enable qualification of predictions, stratification of case difficulty, and principled ensembling of agents~\cite{Kendall_Gal_2017, Ovadia_2019}.

Our dataset uniquely incorporates human expert confidence for all images, providing a rare opportunity to compare model uncertainty with human expert uncertainty. Quantifying uncertainty is complicated by two cardinal aspects of the task. First, uncertainty has no reliable ground truth in real-world settings. Second, uncertainty in the present context contains an aleatoric component--the constitutional discriminability of the classes--and an epistemic component--the agent's ability to discriminate between them. The former is determined by the domain, the latter by the characteristics of the agent. In the ideal case, the epistemic uncertainty is zero, and the agent's uncertainty is wholly aleatoric, determined by how much discriminant signal the data contains. In such a case, the relation between uncertainty and accuracy ought to approximate that of an ideal psychophysical observer detecting a noisy signal, i.e. the uncertainty measure should resemble discriminability.

\begin{figure}[ht]
    \centering
    \includegraphics[width=1.0\linewidth]{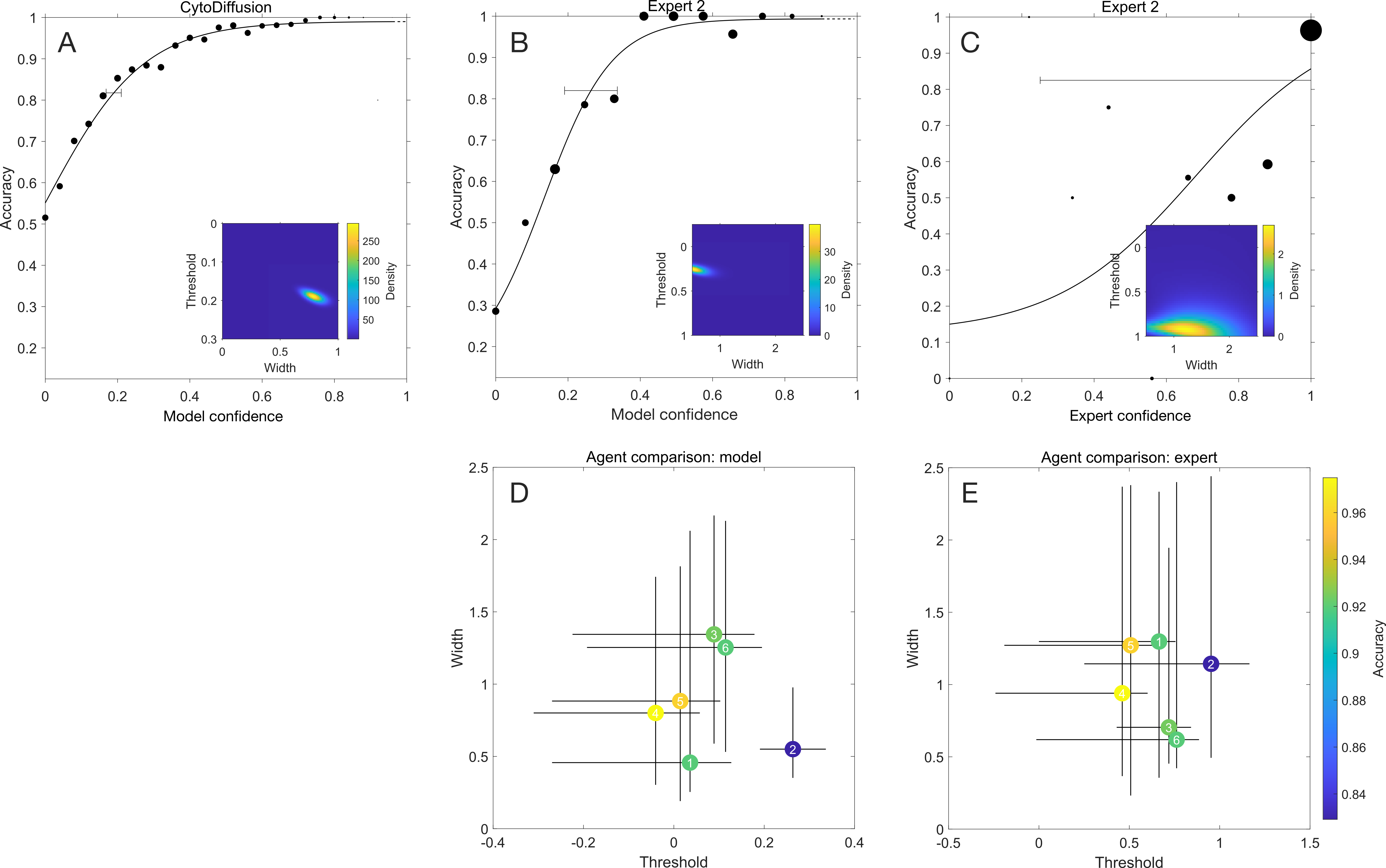}
    \caption{Bayesian psychometric analysis of model and expert performance. A. Psychometric function of CytoDiffusion’s performance on our custom dataset with model confidence as the index of discriminability. The 95 per cent credibility interval of the threshold of the function, estimated at the point of 80 per cent correct unscaled by lapse and guess rate, is shown as an error bar, and the posterior distributions of threshold and width are presented in the inset axes. Note tight bounds on the parameters. B. Psychometric function of the performance of a single human expert, derived from a randomly selected subset of 200 cases for which a consensus ground truth is available, with model confidence as the index of discriminability. C. Psychometric function of the expert shown in B, with consensus expert confidence as the index of discriminability. D \& E. Width and threshold parameter estimates and their 95 per cent credibility intervals for each of the six human experts with model confidence (D) and consensus expert confidence (E) as the indices of signal strength.}
    \label{fig:uncertainty_expert_model}
\end{figure}

This insight allows us to deploy the mature conceptual apparatus of psychometric function estimation to the task of evaluating an agent's uncertainty. In brief, an ideal agent should exhibit a sigmoidal relationship between confidence--the inverse of uncertainty--and accuracy, compactly described by four parameters. The threshold of the function is its location at a chosen accuracy level; the slope is the width or slope of the function; the floor is the chance level, determined by the task; and the ceiling is the lapse level, determined by the error rate where discriminability is maximally easy. Here we employ well-established Bayesian psychometric modelling techniques to derive a psychometric function for CytoDiffusion's performance (Figure ~\ref{fig:uncertainty_expert_model}A), revealing an excellent fit, with tight posterior distributions on the key threshold and width parameters (Figure ~\ref{fig:uncertainty_expert_model}A, inset axes). Although direct measurement is impossible, this suggests CytoDiffusion's uncertainty is dominated by the aleatoric component and its behaviour is close to that of an ideal observer.

This conclusion is reinforced by evaluating individual human expert performance, judged against expert consensus, with CytoDiffusion's confidence as the measure of discriminability. The resultant function, illustrated for Expert 2 (Figure ~\ref{fig:uncertainty_expert_model}B), not only exhibits a good fit, but describes the relationship better than consensus human expert confidence (Figure ~\ref{fig:uncertainty_expert_model}C), suggesting CytoDiffusion's metacognitive abilities are superior to human experts here. Examination of the estimated threshold and width parameters for each human expert, with CytoDiffusion (Figure ~\ref{fig:uncertainty_expert_model}D)) or human expert (Figure ~\ref{fig:uncertainty_expert_model}E) confidence, shows CytoDiffusion's measure can distinguish between the varying abilities of human experts better than they themselves can.

\subsection*{CytoDiffusion Excels at Detecting Anomalous Cell Types} \label{sec:Results_Anomaly_Detection}

For each dataset, we state the anomalous class and the performance, given by the area under the received operating characteristic curve (AUC), of CytoDiffusion in detecting it.
For the Bodzas dataset~\cite{Bodzas_Kodytek_Zidek_2023}, with blasts as the abnormal class, ViT achieved an AUC of 0.698 whereas CytoDiffusion achieved an AUC of 0.982.
Similarly, for the PBC dataset~\cite{Acevedo}, with erythroblasts as abnormal, ViT achieved an AUC of 0.918 whereas CytoDiffusion achieved an AUC of 0.990.
These high AUC values demonstrate our model's capability to distinguish between normal cells it was trained on and abnormal cell types not present in the training data.

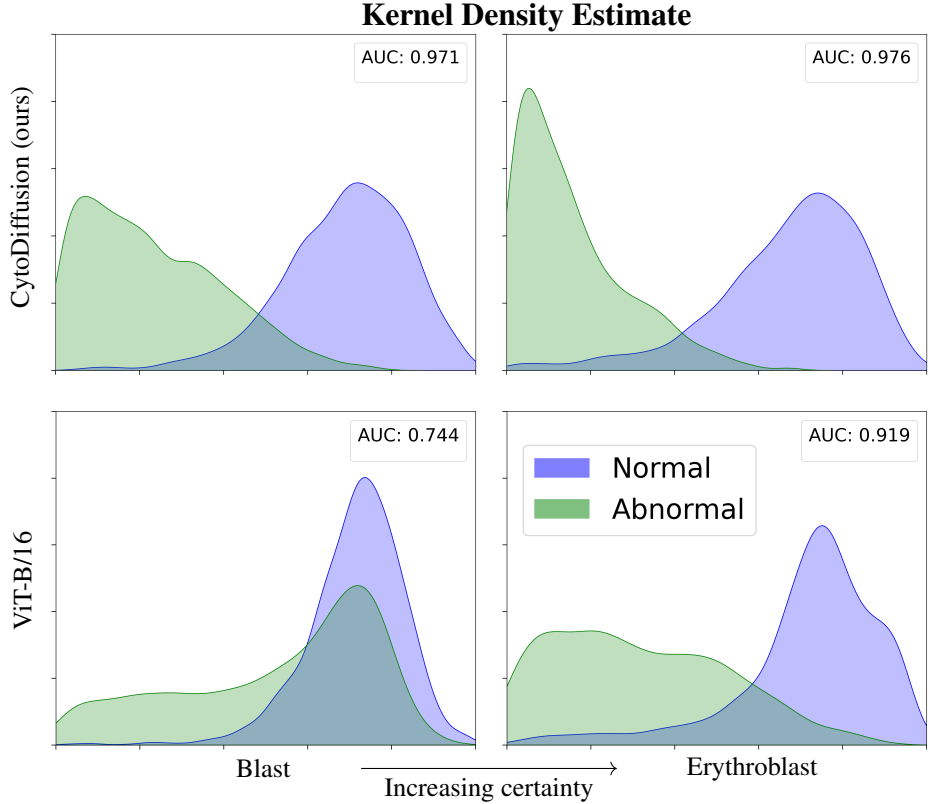
\begin{figure}[ht]
    \centering
    \begin{tikzpicture}
        \node[inner sep=0pt] (fig1) at (0,0) {\includegraphics[width=0.35\textwidth]{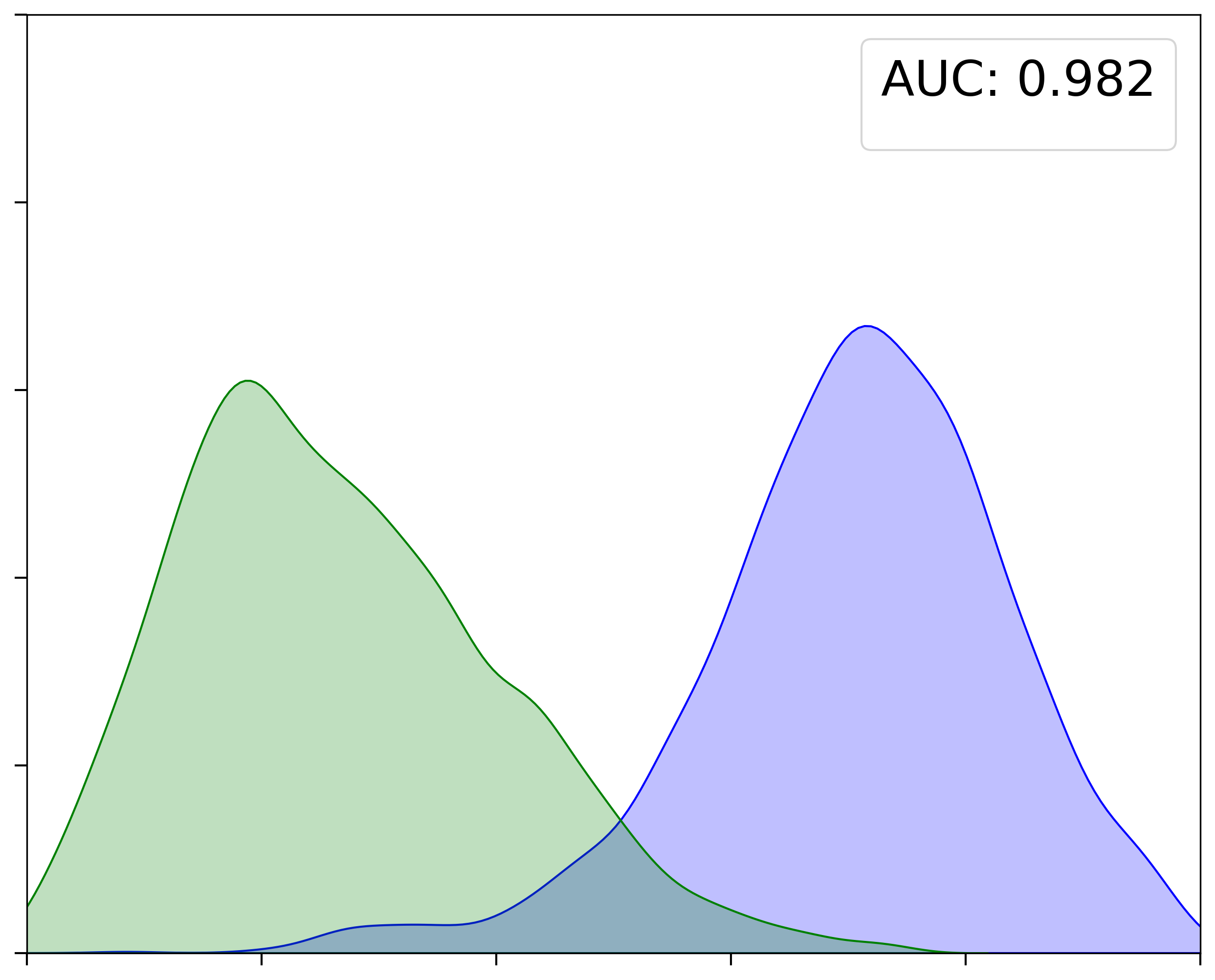}};
        \node[inner sep=0pt] (fig2) at (6,0) {\includegraphics[width=0.35\textwidth]{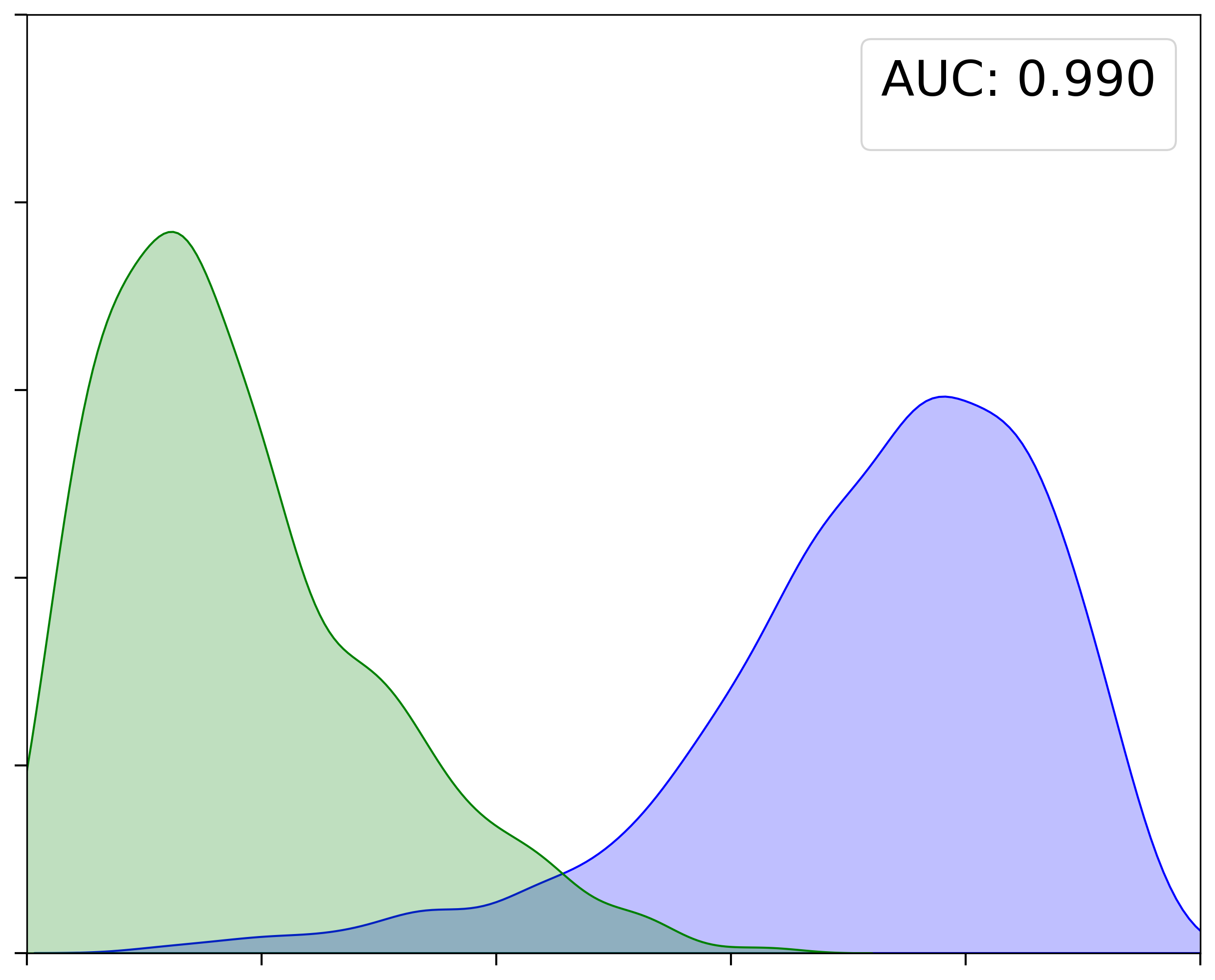}};
        \node[inner sep=0pt] (fig3) at (0,5) {\includegraphics[width=0.35\textwidth]{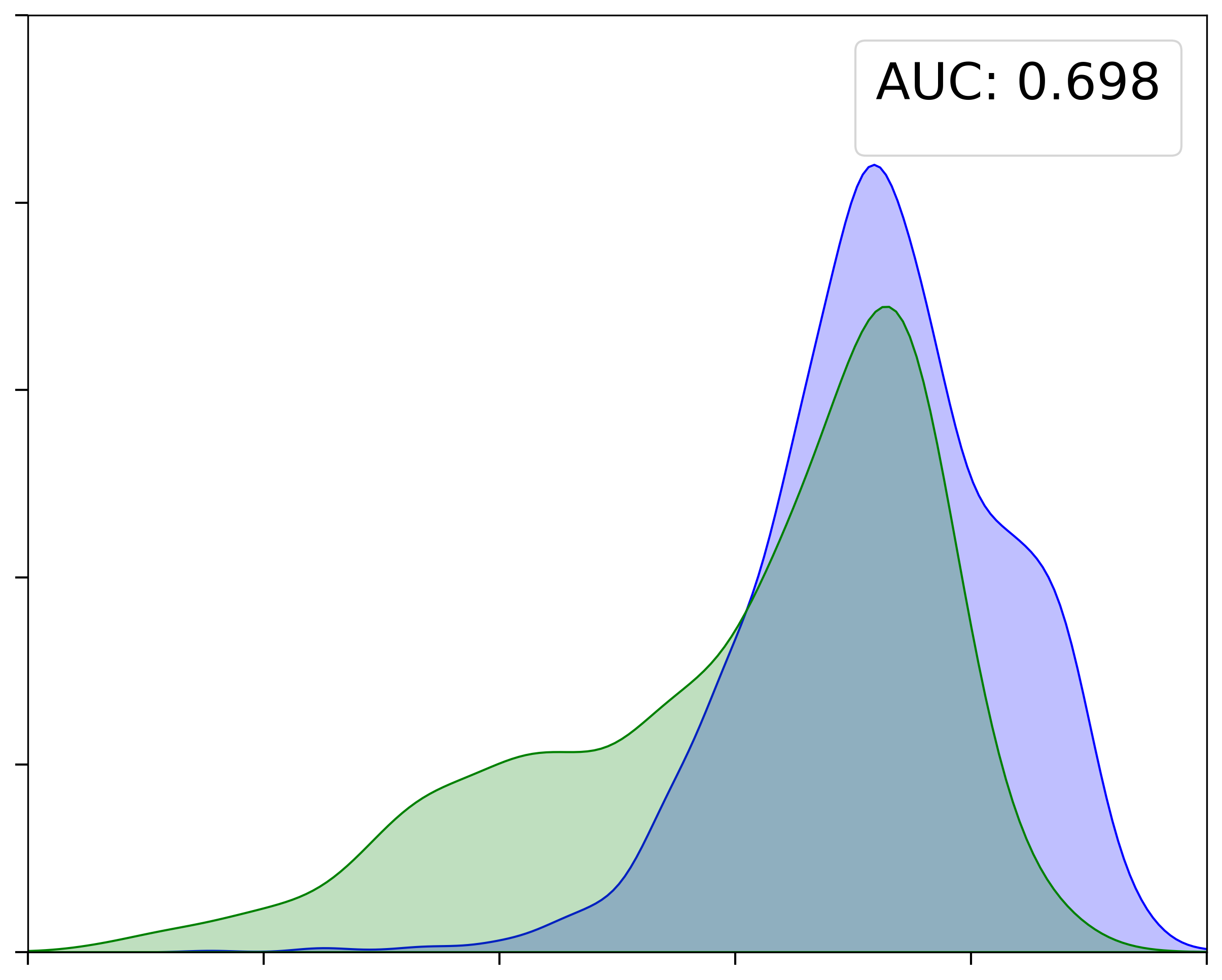}};
        \node[inner sep=0pt] (fig4) at (6,5) {\includegraphics[width=0.35\textwidth]{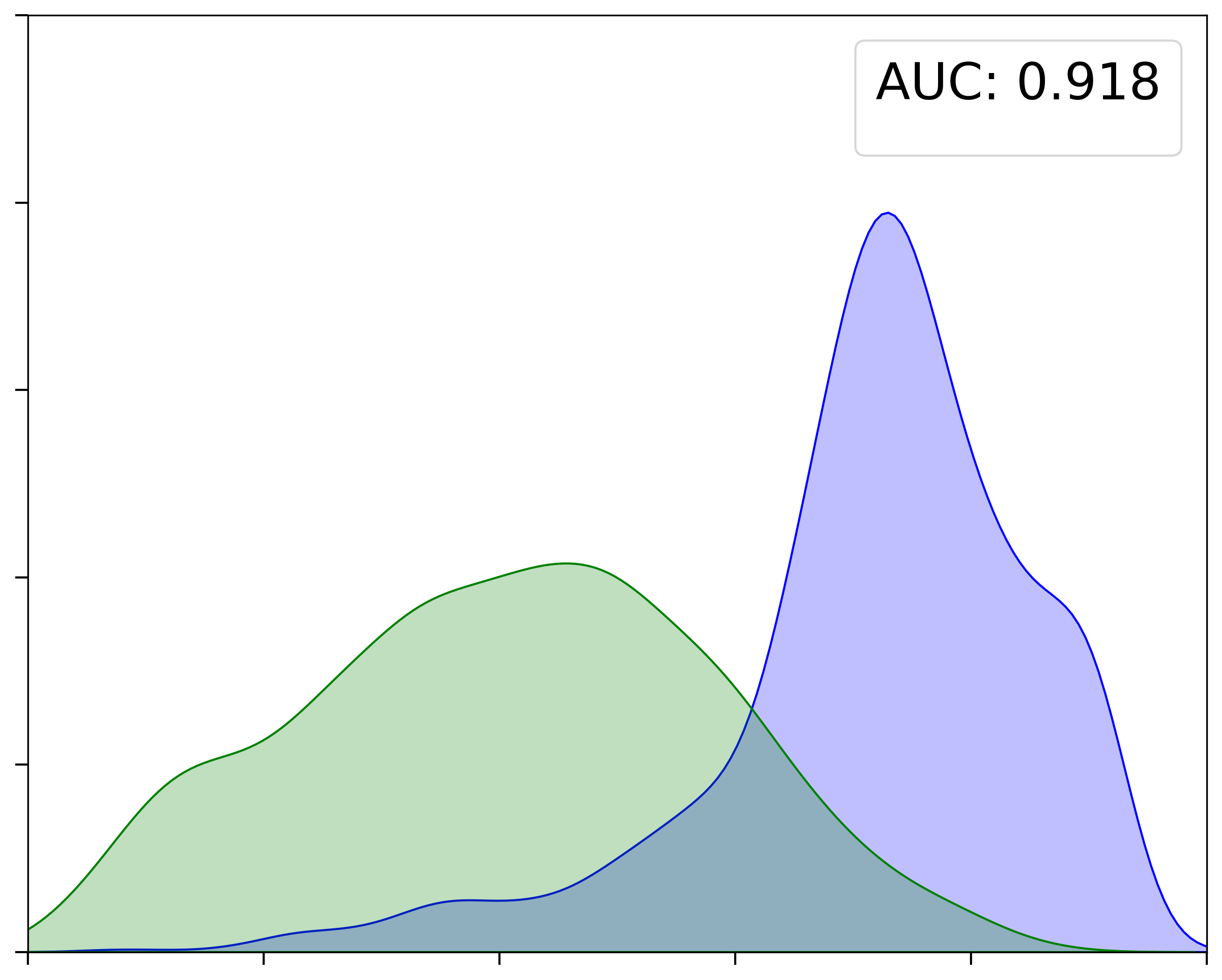}};

        \node[inner sep=0pt] at (3.015,7.85) {\includegraphics[width=0.15\textwidth]{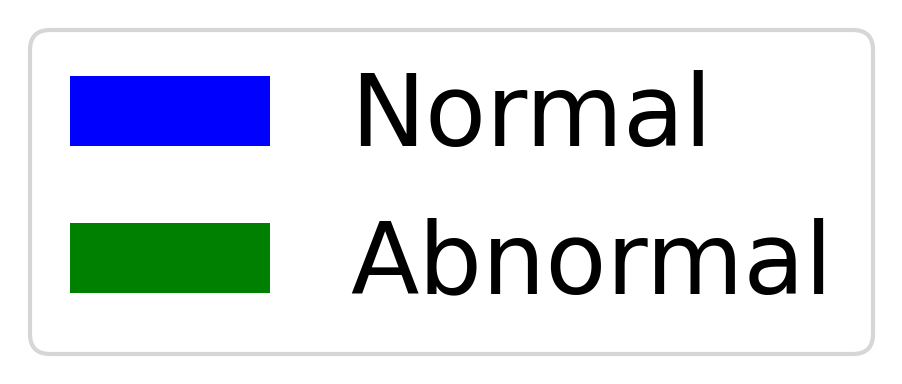}};
        \node[below] at (3,-2.5) {Certainty};

        \node[below] at (-2.75,-2.2) {\scriptsize 0};
        \node[below] at (-1.638,-2.2) {\scriptsize 0.2};
        \node[below] at (-0.526,-2.2) {\scriptsize 0.4};
        \node[below] at (0.586,-2.2) {\scriptsize 0.6};
        \node[below] at (1.698,-2.2) {\scriptsize 0.8};
        \node[below] at (2.81,-2.2) {\scriptsize 1};
        \node[below] at (3.24,-2.2) {\scriptsize 0};
        \node[below] at (4.354,-2.2) {\scriptsize 0.2};
        \node[below] at (5.468,-2.2) {\scriptsize 0.4};
        \node[below] at (6.582,-2.2) {\scriptsize 0.6};
        \node[below] at (7.696,-2.2) {\scriptsize 0.8};
        \node[below] at (8.81,-2.2) {\scriptsize 1};

        \node[below] at (0.03,7.75) {PBC};
        \node[below] at (6.03,7.75) {Bodzas};

    \end{tikzpicture}
        \caption{Kernel density estimate figures comparing the anomaly detection performance of ViT-B/16 (top row) with CytoDiffusion (bottom row) for blasts (left column) and erythroblasts (right column). The horizontal axis represents the certainty measure, normalised to $[0, 1]$ by dividing by the maximum certainty for each model and dataset. The AUC values indicate the model's ability to distinguish between normal and abnormal samples.}
    \label{fig:kde_plots}
\end{figure}

\subsection*{CytoDiffusion Shows Robustness to Domain Shifts}

To assess the generalisability of the CytoDiffusion model, it was trained using the Raabin-WBC dataset, with testing performed on two distinct datasets: Test-B and LISC.
Test-B is part of the Raabin-WBC dataset, but the authors have provided a specific split for testing purposes.
Both Test-B and LISC differ from the training data in camera and microscope types, while LISC additionally uses a different staining technique. In Table~\ref{tab:domain_shift}, we see that CytoDiffusion outperforms the competitor models particularly in LISC where there is a gain of 11.47 in balanced accuracy against the next best model with no overlap of confidence intervals.
\begin{table}[ht]
\centering
\begin{tabular}{lcc}
\toprule
\textbf{Method} & \textbf{Test-B} & \textbf{LISC} \\
\midrule
Li and Liu~\cite{Li_Liu_2024} & $92.36 \pm 0.73$ & $62.51 \pm 2.83$ \\
Li and Liu~\cite{Li_Liu_2024} & $86.84 \pm 1.91$ & $64.96 \pm 2.63$ \\
S. Tsutsui~\cite{Tsutsui_Su_Wen_2023} & $92.98 \pm 1.84$ & $ 70.84 \pm 6.10 $ \\
ViT-B/16~\cite{dosovitskiy2021image} & $96.75 \pm 2.94$ & $62.28 \pm 14.47$ \\
EfficientNetV2-M~\cite{tan2021efficientnetv2} & $96.91 \pm 2.24$ & $74.38 \pm 4.87$ \\
CytoDiffusion (ours) & $\boldsymbol{98.39} \pm 0.34$  & $\boldsymbol{85.85} \pm  2.66$ \\
\bottomrule
\end{tabular}
\vspace{0.1in}
\caption{Performance under domain shift conditions. Models were trained on the Raabin-WBC dataset and evaluated on Test-B and LISC datasets, representing different degrees of domain shift. Balanced accuracy scores (mean $\pm$ standard deviation) from five independent training sessions are reported.} \label{tab:domain_shift}
\end{table}

\subsection*{CytoDiffusion Outperforms Competitors in Low-Data Scenarios} \label{Results Efficiency}

\begin{figure}[ht]
    \centering
    \includegraphics[width=0.8\linewidth]{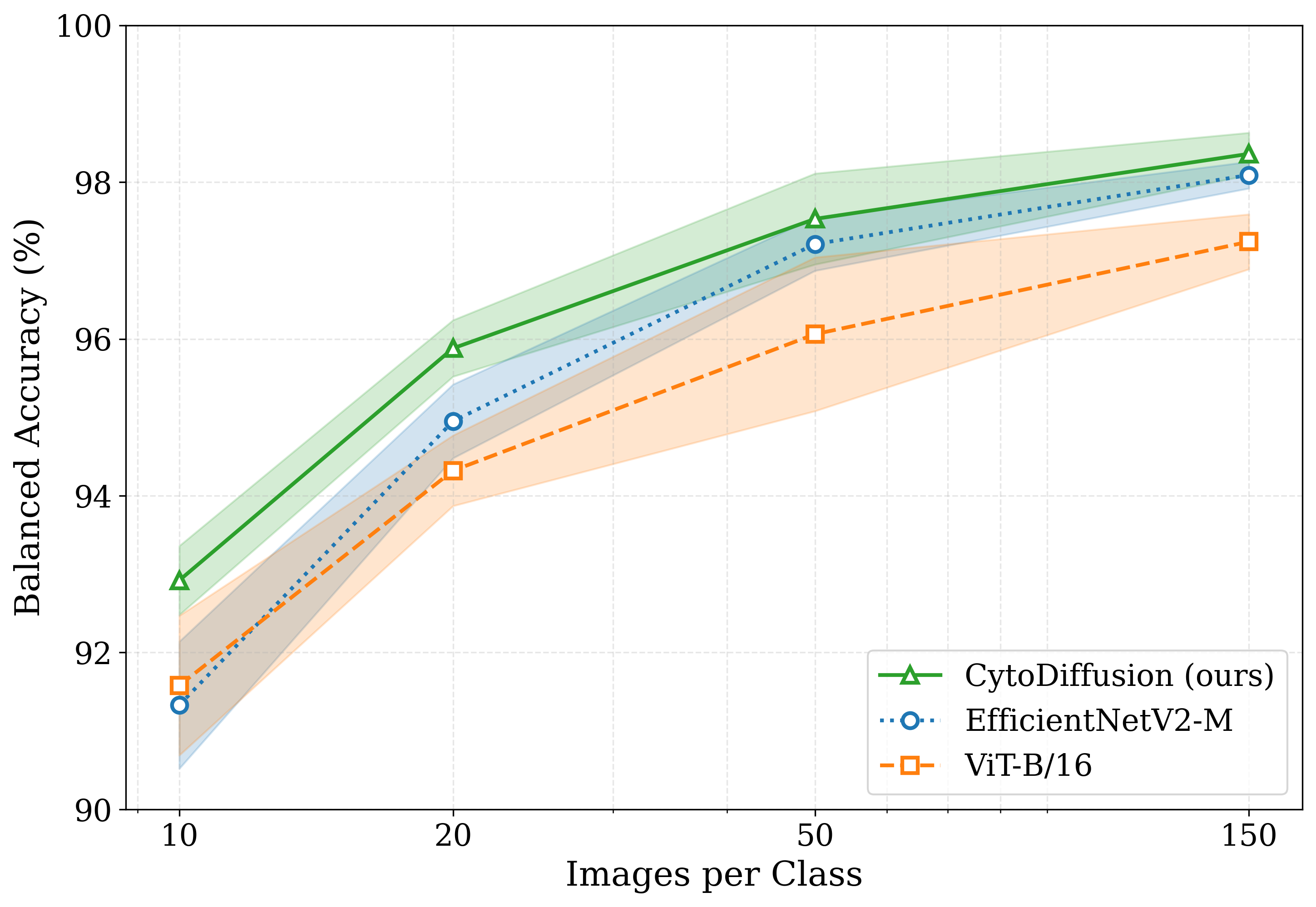}
    \caption{Model performance comparison under low-data conditions. Shaded areas represent the standard deviation from five independent training sessions.}
    \label{fig:efficiency_graph}
\end{figure}

To compare model performances under different training data availability rates, we used the PBC dataset~\cite{Acevedo}. with training carried out with subsets of 10, 20, 50, and 150 images per class, simulating conditions of sparse data availability.
In Figure~\ref{fig:efficiency_graph} we see that CytoDiffusion consistently outperforms competitor discriminative models EfficientNetV2-M and ViT-B/16 across all data availability levels, with the advantage particularly pronounced in the most data-scarce conditions.

\subsection*{CytoDiffusion Provides Visual Explanations Through Counterfactual Heatmaps}

\begin{figure}[htb]
    \centering
    \begin{tikzpicture}
        \node (table) {
            \begin{tabular}{ccccccc}
                 Original & & & & Heatmap & & Overlay  \\[0.05in]
                 \includegraphics[width=0.16\textwidth]{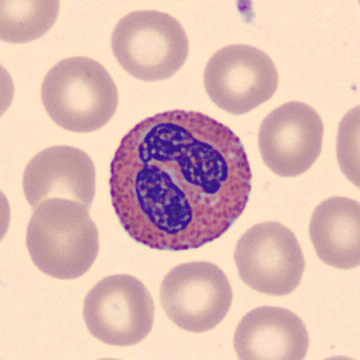}
                 &
                 \raisebox{0.09\textwidth}{%
                 \begin{tikzpicture}
                  \draw[thick, -{Stealth[length=1.5mm, width=1.5mm]}] (0,0) -- (0.5,0);
                \end{tikzpicture}
                }
                &
                 \raisebox{0.055\textwidth}{%
                 \begin{tikzpicture}
                 \definecolor{light_blue}{rgb}{0.847, 0.906, 0.984}
                  \definecolor{blue}{rgb}{0.412, 0.553, 0.741}
                  \node[xshift=-0.5cm, text width=3cm, align=center, fill=light_blue, draw=blue, rounded corners=5pt] (note) {``If this image were a neutrophil, what would you change?''};
                \end{tikzpicture}
                }&
                 \raisebox{0.09\textwidth}{%
                 \begin{tikzpicture}
                  \draw[thick, -{Stealth[length=1.5mm, width=1.5mm]}] (0,0) -- (0.5,0);
                \end{tikzpicture}
                }&
                \includegraphics[width=0.16\textwidth]{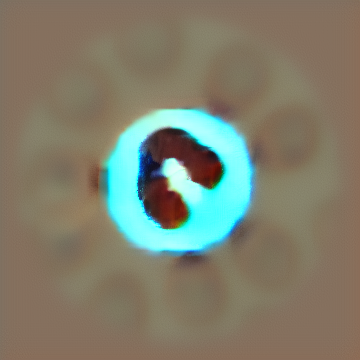}
                &
                \raisebox{0.09\textwidth}{%
                 \begin{tikzpicture}
                  \draw[thick, -{Stealth[length=1.5mm, width=1.5mm]}] (0,0) -- (0.5,0);
                \end{tikzpicture}
                }&
                \includegraphics[width=0.16\textwidth]{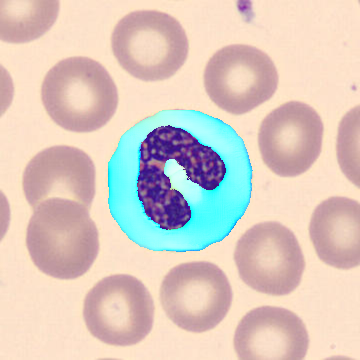}
                \\
            \end{tabular}
        };
    \end{tikzpicture}
    \caption{Counterfactual heatmap visualisation. Left: Original eosinophil image. Centre-right: Counterfactual heatmap ($\boldsymbol{H}_{\text{neutrophil}}$) illustrating the areas that differ when the model considers the image as a neutrophil. (For an example of how a neutrophil can look, see Figure~\ref{fig:heatmap_summary}.) Far right: Overlay of the thresholded heatmap on the original image.}
    \label{fig:counterfactual}
\end{figure}

Counterfactual heatmaps highlight the regions of an image that would need to change for it to be classified as a different cell type.
In Figure~\ref{fig:counterfactual}, we used an eosinophil as an example and prompted the model to consider what alterations would be necessary for this cell to be classified as a neutrophil, generating a heatmap ($\boldsymbol{H}_{\text{neutrophil}}$) that highlights regions where there is large error in the latent space between the two classes. The overlay of this heatmap on the original image reveals that the model focuses primarily on distinguishing granularity between neutrophils and eosinophils, with areas of significant colour deviation from the background indicating the most critical regions of difference.

\begin{figure}[ht]
    \centering
    \begin{tikzpicture}
        \pgfmathsetmacro{\imagewidth}{0.86}
        \pgfmathsetmacro{\labelspace}{0.03}

        \node[anchor=south west,inner sep=0] (image) at (\labelspace,0) {\includegraphics[width=\imagewidth\textwidth]{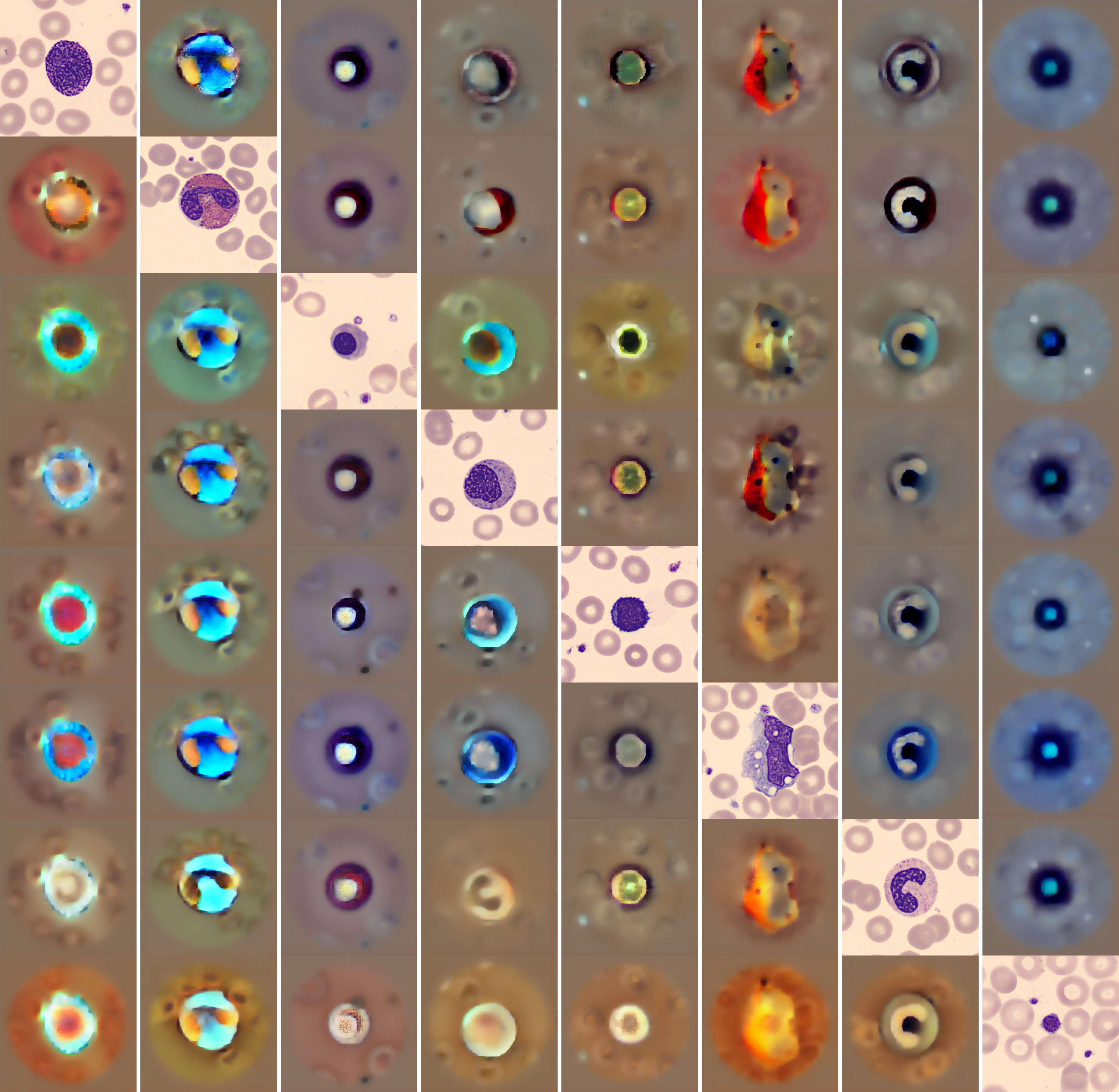}};

        \begin{scope}[x={(image.south east)},y={(image.north west)}]
            \foreach \x/\label in {0.1/\strut Basophil,1.05/\strut Eosinophil,2.05/ \strut Erythroblast,3/\strut {Imm.\ Gr.},3.96/\strut Lymphocyte,4.95/\strut Monocyte,5.93/\strut Neutrophil,6.9/\strut Platelet}
            {
                \pgfmathsetmacro{\xpos}{(\x/7 * 0.90) + 0.05}
                \node[anchor=south,rotate=0,align=center] at (\xpos, 1.00) {\footnotesize\label};
            }

            \foreach \y/\label in {0.9375/Basophil,0.8125/Eosinophil,0.6875/Erythroblast,0.5625/{\shortstack[l]{Immature\\Granulocyte}},0.4375/Lymphocyte,0.3125/Monocyte,0.1875/Neutrophil,0.0625/Platelet}
            {
                \node[anchor=south,rotate=60,align=left] at (-0.02,\y) {\footnotesize\label};
            }

            \draw[<->, thick, blue] (0.4, 1.05) -- (0.6, 1.05) node[midway, above, font=\small\color{blue}] {Source cell type};
            \draw[<->, thick, blue] (-0.11, 0.4) -- (-0.11, 0.6) node[midway, above, sloped, font=\small\color{blue}] {Target cell type};
        \end{scope}
    \end{tikzpicture}
    \caption{Counterfactual heatmaps for cell type transitions in the PBC dataset. The diagonal shows original images of each cell type, serving as the source. Each off-diagonal element in the same column represents a counterfactual heatmap ($\boldsymbol{H}_c$) showing the transition from the diagonal element (source) to the cell type of that row (target). Areas in the heatmap with colours that deviate significantly from the background indicate regions where there is large error in the latent space between the two classes.}
    \label{fig:heatmap_summary}
\end{figure}
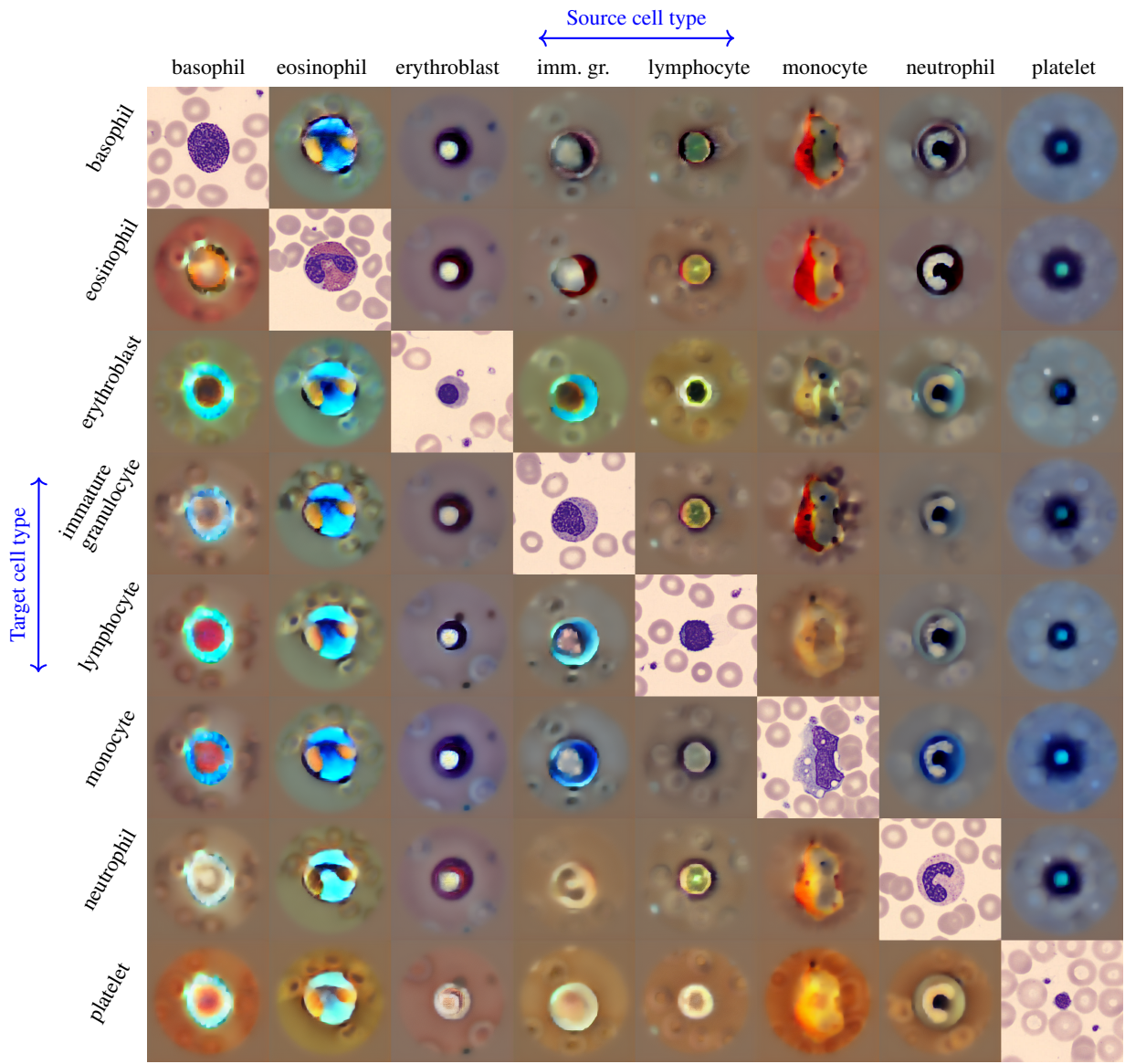

To provide a comprehensive view of CytoDiffusion's abilities across all cell types, in Figure \ref{fig:heatmap_summary}) we generated counterfactual heatmaps for each possible class transition in the PBC dataset. This visualisation offers insights into the model's decision-making process for each cell type. For instance, when considering the transition from neutrophil to eosinophil (row 2, column 7), the model highlights regions in the cytoplasm (darker areas) where features should be added, while largely maintaining the nuclear shape.

Notably, the heatmaps also reveal the model's understanding of subtle differences between similar cell types. In the transition from monocyte to immature granulocyte (Figure \ref{fig:heatmap_summary}, row 4, column 6), the model indicates the difference in cytoplasm between the more acidophilic cytoplasm of the immature granulocytes in comparison to the greyish-blue monocytic cytoplasm. Intriguingly, the model also suggests the filling of the monocytic vacuoles (appearing as dark spots in the heatmap). This captures one of the typical morphological findings in monocytes differentiating them from other normal blood cells and demonstrates the model's ability to focus on nuanced information.

\section*{Discussion}
This study introduces not only the novel CytoDiffusion model for haematological cell image classification but also a principled evaluative framework.
Our approach is motivated by the desire to achieve a model with superhuman fidelity, flexibility and metacognitive awareness which can capture the distribution of all possible morphological appearances.
These ambitions are attainable only by eliciting structure in the data beyond that which can be obtained simply from expert labelling of images.
Both our methodology and evaluative framework are grounded in the recognition of the inherent complexity of the target system along with the inherent constraints for AI-modelling with medical imaging data, e.g. relatively small dataset regimes and instrumental variation reflected in images.
In addition, we argue that medical imaging models demand a holistic evaluation framework that includes not only simple performance evaluation but also the model's (a) robustness to domain shift, (b) ability to detect anomalies and (c) performance in low data availability scenarios, (d) ability to quantify its uncertainty accurately and (e) ability to provide interpretable outputs describing the decision making process.

Robustness of a classification model to domain shift, i.e. its ability to generalise across different imaging conditions, is crucial for its practical application in clinical settings. This generalisation capability is particularly important in haematology, where variations in microscope types, camera systems, and staining techniques are common across different laboratories and hospitals. A model that performs well only under specific imaging conditions would have limited utility in real-world clinical practice, potentially leading to inconsistent or unreliable diagnoses when deployed in new environments.
Additionally, the ability to identify rare or unexpected cell types is crucial in clinical scenarios, where the detection of abnormal cells can have significant diagnostic implications. We propose that future models report performance scores for similar anomaly detection tasks, providing a standardised approach for assessing a model's ability to flag potentially abnormal findings in medical imaging.
Finally, the ability of a model to perform well with limited training data is crucial in medical imaging contexts, where large, well-annotated datasets are often challenging to obtain. We propose that new models in this domain should be evaluated on their performance in low-data regimes to assess their efficiency.
Our results demonstrate the failures of discriminative models, even state-of-the-art architectures trained on large-scale datasets, when subjected to this more holistic evaluative framework whilst CytoDiffusion demonstrates substantial improvements, without relying on careful hyperparameter tuning or data-specific architectural refinements.

Additionally, in any classification task, whether performed by a human or an ML algorithm, it is highly informative to understand the uncertainty in the final decision, a surrogate for how difficult the sample is to classify. This is particularly true for clinical data where the classification result can inform an intervention or treatment decision.
Currently, a method for optimal evaluation of model uncertainty is not established in the ML literature. We introduce a novel framework for evaluating an agent's uncertainty--human or machine--based on the expected structure of the purely aleatoric uncertainty of an ideal psychophysical observer. This approach allows us to quantify model uncertainty as departure from the ideal psychometric function relating purely aleatoric uncertainty to fidelity. We demonstrate that CytoDiffusion produces superior estimates of uncertainty than human experts themselves.

Moreover, a key requirement of deploying machine learning models in clinical scenarios is their ability to provide interpretable results. CytoDiffusion addresses this need through the generation of counterfactual heatmaps, which highlight the regions of an image that would need to change for it to be classified as a different cell type, giving valuable insights into CytoDiffusion's decision-making process, offering a detailed visualisation of the morphological distinctions the model identifies between cell types. This granular view of the model's reasoning enhances our understanding of its classificatory mechanisms, potentially contributing to the development of more transparent and interpretable models across various domains of medical image analysis.
A distinctive feature of our approach is that these heatmaps are the direct output of our model, requiring no post-processing or additional colour mapping. This direct output capability preserves the integrity of the model's decision-making rationale, offering an unfiltered view of its reasoning process.

Our approach is not without limitations. The inference process of CytoDiffusion is computationally expensive, scaling with the number of classes. However, this is less problematic in the medical domain, where datasets typically have far fewer classes than general image classification tasks like ImageNet. Moreover, CytoDiffusion's adaptive allocation of computing resources, dedicating more effort to challenging images, makes it more effective.

There are many possible extensions to this study. Firstly, we have not fully exploited the representation learning capabilities of generative models, where there is significant potential for further advances, possibly through e.g. the use of generative modelling architectures with compact latents.
Additionally, generative models also enable the assurance of equity in medical diagnostics. Although not implemented in this study, conditioning on minority characteristics would allow for legibility of differences in the model's perception across diverse demographic groups. This capability is crucial for both detecting potential biases and enabling targeted remedial actions, ensuring fair and equitable application of AI in healthcare.

In conclusion, this work establishes a new paradigm for medical image analysis, particularly in haematology. By embracing the inherent complexities of biological systems, and the limitations of current data regimes, our generative approach with CytoDiffusion and comprehensive evaluation framework pave the way for more robust, interpretable, and equitable AI systems in healthcare. Future work should focus on refining these methods, exploring their applicability to other medical imaging domains, and further enhancing their computational efficiency without compromising performance.

\section*{Methods}

\subsection*{Diffusion Classifiers}

Recent studies have shown that diffusion models can also be utilised as classifiers~\cite{Chen_Dong_Wang_Yang_Duan_Su_Zhu_2024, Clark_Jaini_2023, Li_Prabhudesai_Duggal_Brown_Pathak_2023, Generative_Classifiers}. We use a latent diffusion model as described in Supplementary Material~\ref{latent_diffusion_setup}. Given an input image $\boldsymbol{x}$, we want to predict the most probable class $\hat{c}$. This can be formalised as finding the class $\hat{c}$ that maximises the posterior probability $p(c = c_k | \boldsymbol{x})$. Using Bayes' theorem, this is equivalent to
\begin{equation}
\hat{c} = \argmax_{c_k}\, p(c = c_k | \boldsymbol{x}) = \argmax_{c_k} \, p(\boldsymbol{x} | c =
 c_k) \cdot p(c = c_k) = \argmax_{c_k} \log p(\boldsymbol{x} | c = c_k) \, ,
\end{equation}
assuming a uniform prior over the classes, $p(c = c_k) = \frac{1}{K}$, where $K$ is the number of classes.

As we do not have direct access to the (negative) log likelihood, we use our loss function to approximate it instead, and therefore take
\begin{equation}
\hat c = \argmin_c \mathbb{E}_{\boldsymbol{\epsilon}, t} \left[ w_t \left \| \boldsymbol{\epsilon} - \boldsymbol{\epsilon}_\theta(\boldsymbol{z}_t, t, c) \right \|^2_2 \right],
\end{equation}
with the weight $w_t$ discussed in Supplementary Material~\ref{latent_diffusion_setup}.  To achieve this, we randomly pick a time step $t$ and noise $\boldsymbol{\epsilon}$ and calculate the error
\begin{equation}
\left \| \boldsymbol{\epsilon} - \boldsymbol{\epsilon}_\theta(\boldsymbol{z}_t, t, c) \right \|^2_2 \, ,
\end{equation}
for all class labels $c$. These error values are then normalised using $w_t$ and the results are stored.
The process then repeats with newly sampled values of $t$ and $\boldsymbol{\epsilon}$.
For an intuitive explanation of how our model makes its predictions, we refer to Supplementary Figure~\ref{fig:kitten}.

Using a procedure similar to~\cite{Clark_Jaini_2023}, we repeatedly gather new sets of errors for each candidate class. Classes that are less likely to have the lowest-error are progressively eliminated. This is achieved using a paired Student's $t$-test, which compares the errors across classes. However, given that the errors do not perfectly follow the standard assumptions of a Student's $t$-test, we employ a small p-value of $2\times10^{-3}$ to ensure robustness. This iterative procedure continues until there is only one class left or a maximum number of iterations is reached. Unless stated otherwise, we use a minimum of 20 iterations and a maximum of 2000 iterations in our experiments. For an analysis of different p-values tested, as well as various minimum and maximum numbers of iterations, we refer to Supplementary Figure~\ref{fig:ablation_pruning}.

\subsection*{General Training Setup}

Unless otherwise specified, we employed the following training configuration for all experiments. We used Stable Diffusion 1.5~\cite{Rombach_Blattmann_Lorenz_Esser_Ommer_2022} as our base model. For class conditioning, we bypassed the tokeniser and text encoder, directly feeding the model with one-hot encoded vectors for each class, replicated vertically and padded horizontally to match the expected matrix of $77 \times 768$ dimensions. We utilised a batch size of 10, a learning rate of $10^{-5}$ with linear warmup over 1000 steps, and trained on an A100-80GB GPU. Details of the training and inference parameters are in the Supplementary Materials~\ref{train_inf_setup}.

\subsection*{Datasets} \label{Datasets}
We utilised multiple datasets, described in Table~\ref{tab:datasets}, including four which are publicly available and one custom dataset, CytoData, to develop and evaluate our diffusion classifier for haematological cell image classification. CytoData, available at \textbf{[insert upon publication]}, is an anonymised dataset consisting of 559,808 single cell images from 2,904 blood smear slides obtained from Addenbrooke's Hospital in Cambridge, UK, with a labelled subset of 5,000 images across 10 classes. These images were created using CellaVision, a specialised imaging technology for cellular analysis. The labelling strategy is described in the Supplementary Materials~\ref{cytodata_labelling}.
Notably, when labelling CytoData, we included an artefact class, addressing a critical challenge in clinical applications, as blood smear slides often contain artefacts which may be mistaken for cells by deep learning-based cell detection models. By explicitly modelling these artefacts, CytoData aims to enhance clinical applicability.
A distinctive feature of CytoData is the inclusion of labeller confidence scores. This addition provides valuable information for various analyses beyond simple correlations.

\begin{table}[h]
    \centering
    \begin{threeparttable}
        \caption{Summary of the datasets used in our experiments.}
        \label{tab:datasets}
        \begin{tabular}{lcccc}
            \toprule
            \textbf{Dataset} & \textbf{\# Images} & \textbf{\# Classes} & \textbf{Labeller Confidence} & \textbf{Artefacts} \\
            \midrule
            Raabin-WBC~\cite{Raabin_WBC} & 16,633\tnote{$\dag$} &  5 & \ding{55} & \ding{55} \\
            LISC~\cite{LISC} &  257 & 5 & \ding{55} & \ding{55} \\
            PBC~\cite{Acevedo} & 17,092 & 8 & \ding{55} & \ding{55} \\
            Bodzas~\cite{Bodzas_Kodytek_Zidek_2023} & 16,027 & 9 & \ding{55} & \ding{55} \\
            CytoData (ours) & 5,000 & $9+1$\tnote{*} & \checkmark & \checkmark \\
            \bottomrule
        \end{tabular}
        \begin{tablenotes}
            \item[$\dag$] Split into training (10,175 images), Test-A (4,339 images), and Test-B (2,119 images) sets.
            \item[*] 9 regular classes of blood cells plus 1 artefact class.
        \end{tablenotes}
    \end{threeparttable}
\end{table}

\subsection*{Authenticity Test}

To assess the quality and authenticity of our fine-tuned diffusion model's synthetic blood cell images, we conducted an authenticity test with expert haematologists. This evaluation was designed to determine whether the model could effectively capture the underlying distribution of blood cell images across various cell types, a capability that traditional discriminative models are not inherently required to possess. Additionally, we sought to assess the accuracy of the generated cell types. Further details are in Supplementary Materials~\ref{turing_test_setup}.
Ten haematology specialists from our research group, with \{34, 28, 25, 15, 10, 9, 6, 5, 5, 1\} years of experience in blood microscopy, participated in the authenticity test. Participants were informed that half of the images presented to them would be real images from our dataset, while the other half would be synthetic images generated by our model. Each specialist was presented with the 288 images in a randomised order and asked to perform two tasks: (1) identify whether each image was synthetic or real and (2) classify each image into one of the nine designated blood cell types.

\subsection*{In-Domain Performance} \label{In-Domain Performance}
To establish a baseline for our diffusion-based classifier, CytoDiffusion, we evaluated its performance on standard in-domain classification tasks using three datasets: our custom dataset, Raabin-WBC, and PBC. For the Raabin-WBC dataset, we utilised the predefined train split (90\% for training, 10\% for validation) and tested on the predefined Test-A dataset. For the PBC dataset, we employed an 80--10--10 split for train--validation--test.
For our custom dataset, we employed a cross-validation approach where we divided the dataset into four folds and fine-tuned four separate models. For each fold used in training, we further split the data into 80\% for training and 20\% for validation and we trained each model for 60,000 iterations.
For the Raabin-WBC and PBC datasets, we trained our model for 72,000 steps. To provide a basis for comparison, we also trained and evaluated EfficientNetV2-M and ViT-B/16 models.
Given the significant class imbalances present in these datasets (Raabin-WBC Test-A ranges from 89 to 2,660 images per class, our custom dataset from 103 to 1,425, and PBC from 1,199 to 3,314), we report both overall accuracy and balanced accuracy.
It is important to note that we have excluded some studies from our comparison due to methodological differences that could lead to unfair or misleading comparisons. Specifically, we are not comparing with papers that do not have a conventional train--validation--test split \cite{Zhang_Han__2022, Long_Peng_Song_Xia_Sang_2021}, or those that do not test on the predefined test set \cite{Firat_2024, Rubin_Anzar_2023}.

\subsection*{Uncertainty Measure} \label{Uncertainty Measure}

To evaluate the quality of CytoDiffusion's uncertainty measure, we exploited the decomposability of uncertainty into model and aleatoric components. An ideal model, indeed any ideal agent, should contribute no uncertainty of its own, leaving aleatoric uncertainty as the sole residue. If so, then the uncertainty measure should reduce to the magnitude of the discriminative signal, and the relation between the uncertainty measure and model fidelity should conform to that of an ideal observer of a noisy signal. This allows us to exploit psychometric function modelling to quantify how close an agent's uncertainty--machine or human--is to the aleatoric floor~\cite{Gescheider_1997}.
We utilised the cross-validated models trained on our custom dataset, as described in the ``\nameref{In-Domain Performance}'' section.
For quantifying the model's uncertainty, we calculated the difference between the two classes with the smallest error, rescaled in the interval $[0, 1]$. To establish a measure of labeller uncertainty, we mapped the confidence levels provided by our expert haematologists to numerical values: 1.0 for High Confidence, 2/3 for Moderate Confidence, 1/3 for Low Confidence, and 0.0 for No Confidence. For images assessed by multiple labellers, we averaged their uncertainty scores.

Psychometric functions describe the relation between the performance of an observer and a (typically scalar) property of the observed~\cite{schutt2016painfree}. The performance of interest is usually detection or classification, expressed as a function of signal strength on a monotonically increasing scale. Since an ideal model is as confident as the data allows, exhibiting purely aleatoric uncertainty, we can quantify the proximity of a model to that ideal by fitting a psychometric function with model confidence as the index of signal strength. A good measure of uncertainty should conform closely to an ideal observer, yielding a sigmoid curve rising from a chance guess rate, $\gamma$, where uncertainty is maximal, to a lapse rate $\lambda$, where uncertainty is minimal and any errors are not explicable by insufficient information. Further details are in the Supplementary Material~\ref{psycho_analysis}. For $\gamma$, we fix the parameter at 1/10, reflecting the 10 possible classes, and for $\lambda$ we use a beta-distribution with parameters (1,10). We report the threshold and width estimated parameters, and their posterior distributions, citing 95\% credibility intervals.

\subsection*{Anomaly Detection} \label{Methods Anomaly Detection}
To evaluate our model's capability for detecting anomalous cell types, we designed an experiment that simulates real-world scenarios where rare or previously unseen cell types might appear in clinical samples. This approach involved excluding specific abnormal cell classes during training and assessing the model's ability to identify these classes during testing.
We utilised two datasets for this experiment: the Bodzas dataset and the PBC dataset. For the Bodzas dataset, we merged the neutrophil segment and neutrophil band classes into a single neutrophil class and excluded the blasts class, which comprised both lymphoblasts and myeloblasts (5,036 images in total). From the PBC dataset, we excluded the erythroblast class (1,513 images). The data was split into 70\% training, 10\% validation, and 20\% test sets, with all abnormal cells (blasts and erythroblasts) moved to their respective test sets.
We employ an anomaly score inspired by~\cite{MAH201451, rieck2006detecting} to quantify the model's confidence (details in Supplementary Material~\ref{anomaly_details}).
To visualise each model's ability to distinguish between normal and abnormal cells, we generated KDE curves of the confidence differences for both groups. To quantify this ability, we calculated the AUC for each model and dataset.

\subsection*{Domain Shift} \label{Domain Shift}

To evaluate our model's robustness under domain shift conditions, we leveraged the Raabin-WBC dataset~\cite{Raabin_WBC} and the LISC dataset~\cite{LISC}, following methodologies outlined in previous studies~\cite{Li_Liu_2024, Tsutsui_Su_Wen_2023}. We utilised the predefined train split for training (90\%) and validation (10\%), and the Test-B split as one of our test sets. The Test-B split was created using a different microscope and camera type compared to the training and validation set, introducing a domain shift. Additionally, we employed the LISC dataset as a second test set, which was created using different microscope and camera types, as well as a different staining method, further increasing the domain shift challenge. To maintain consistency with previous studies~\cite{Li_Liu_2024, Tsutsui_Su_Wen_2023} and accommodate the lower resolution of LISC images, we resized all images to $224 \times 224$ pixels.
We used a batch size of 32 and trained for 22,000 steps on an NVIDIA RTX A5000 GPU.
For comparison, we also fine-tuned and tested EfficientNetV2-M and ViT-B/16.
To ensure reliable performance evaluation, we repeated the training and evaluation process five times for each of the models (our diffusion-based model, EfficientNetV2-M, and ViT-B/16).

\subsection*{Efficiency in Low-Data Regimes} \label{Efficiency}
To assess the efficiency of our model in learning from a small number of images per class, we designed an experiment using the PBC dataset~\cite{Acevedo}. We randomly selected 250 images per class for the training set, 836 images per class for the test set, and the remaining images for the validation set. From the training set, we randomly sampled subsets of 10, 20, 50, or 150 images per class to simulate low-data scenarios.
We trained CytoDiffusion on an NVIDIA RTX A5000 GPU.
For subsets of 10, 20, 50, and 150 images per class, we trained the model for 20,000, 40,000, 100,000, and 240,000 steps, and saved checkpoints every 1,000, 3,000, 7,000, and 20,000 steps, respectively.
For each subset, we selected the checkpoint with the highest validation accuracy for testing.
For comparison, we also trained and evaluated EfficientNetV2-M and ViT-B/16 models using the same data subsets.
To account for variability in the subsampling and initialisations, we repeated each experiment five times. This involved resampling the specified number of images per class (10, 20, 50, and 150) and retraining all models for each repetition.

\subsection*{Explainability}
One of the integral aspects of CytoDiffusion's utility in clinical settings is its ability to provide explainable predictions. To achieve this, we employ a counterfactual heatmap approach, which elucidates what changes would be necessary for an image to be classified under a different specified class. This method is particularly beneficial for understanding model decisions in complex medical imaging tasks such as the classification of blood cell types.
Initially, we calculate the difference between the original noise, $\boldsymbol{\epsilon}$, and the noise predicted by the model for each class condition $c$. This difference is recorded for all iterations, and the mean error for each condition $c$ is computed as
$\Delta_{c} = \frac{1}{N} \sum_{n=1}^{N} \left(\boldsymbol{\epsilon}_n - \boldsymbol{\epsilon}_\theta(\boldsymbol{z}
_{t_n}, t_n, c)\right)\,$ ,
where $N$ is the number of iterations.
Subsequently, for each class condition $c$, we calculate the deviation from the condition with the minimum error, designated as $\Delta_{\hat c}$, where $\hat c$ is the predicted class. The adjusted $\delta_{c}$ is then $\delta_{c} = \Delta_{c} - \Delta_{\hat c}\,$.
Finally, $\delta_{c}$ are decoded back to the pixel space using the VAE decoder to obtain the counterfactual heatmaps $\boldsymbol{H}_{c} = \mathcal{D}(\delta_{c})\,$, where $\mathcal{D}$ denotes the VAE decoder, and $\boldsymbol{H}_{c}$ represents the heatmap for condition $c$. These heatmaps visually represent modifications that would shift the image classification from the predicted class to the target class $c$, thereby providing a powerful tool for explaining and validating model predictions.\\

\noindent
\textbf{Competing interests.} The authors declare no competing interests. \\

\noindent
\textbf{Data availability.} CytoData is available at \textbf{[insert upon publication]}. All other datasets used in the analysis are publicly available as referenced.\\

\noindent
\textbf{Code availability.} All code and trained models are available at \url{https://github.com/CambridgeCIA/CytoDiffusion}.\\

\noindent
\textbf{BloodCounts! Consortium.} Martijn Schut, Folkert Asselbergs, Sujoy Kar, Suthesh Sivapalaratnam, Sophie Williams, Mickey Koh, Yvonne Henskens, Bart de Wit, Umberto D'Alessandro, Bubacarr Bah, Ousman Secka, Parashkev Nachev, Rajeev Gupta, Sara Trompeter, Nancy Boeckx, Christine van Laer, Gordon A. Awandare, Kwabena Sarpong, Lucas Amenga-Etego, Mathie Leers, Mirelle Huijskens, Samuel McDermott, Willem H. Ouwehand, James Rudd, Carola-Bibiane Schönlieb, Nicholas Gleadall and Michael Roberts.\\

\bibliographystyle{unsrt}
\bibliography{references}
\appendix

\newpage

\section{Latent Diffusion Models Setup} \label{latent_diffusion_setup}
Generative models have seen significant advancements in recent years, with diffusion models emerging as a leading approach for creating high-quality data samples~\cite{Rombach_Blattmann_Lorenz_Esser_Ommer_2022, Chen_Mei_Fan_Wang_2024, Ho_Jain_Abbeel_2020, Song_Sohl-Dickstein_Kingma_Kumar_Ermon_Poole_2021, Song_Ermon_2020}. These models draw inspiration from thermodynamic diffusion processes~\cite{Sohl-Dickstein_Weiss_Maheswaranathan_Ganguli_2015} and have demonstrated impressive results in various tasks, such as image generation, inpainting, and super-resolution~\cite{Saharia_et_al._2022, Rombach_Blattmann_Lorenz_Esser_Ommer_2022, Saharia_Chan_Chang_Lee_Ho_Salimans_Fleet_Norouzi_2022, Lugmayr_Danelljan_Romero_Yu_Timofte_Van_Gool_2022, Saharia_Ho_Chan_Salimans_Fleet_Norouzi_2021}.
Diffusion models operate on the principle of defining a forward diffusion process, which gradually adds noise to the data, converting it into a noise-like distribution. The model then learns a reverse process to denoise the data, effectively reconstructing the original data distribution~\cite{Sohl-Dickstein_Weiss_Maheswaranathan_Ganguli_2015, Ho_Jain_Abbeel_2020}.
In latent diffusion models, data $\boldsymbol{x}$ is first encoded into a latent space using an encoder, typically a Variational Autoencoder (VAE), resulting in latent variables $\boldsymbol{z} = \mathcal{E}(\boldsymbol{x})$ with $\boldsymbol{z}\sim q(\boldsymbol{z})$~\cite{Oord_Vinyals_Kavukcuoglu_2018, Rombach_Blattmann_Lorenz_Esser_Ommer_2022, Kingma_Welling_2022}. The forward and reverse diffusion processes are subsequently applied to these latent variables.

\subsection{Forward Diffusion Process}

The forward diffusion process is formulated as a Markov chain, wherein Gaussian noise is incrementally added to the latent variables over $T$ steps. Let $\boldsymbol{z}_0$ denote a variable from the encoded (latent) data distribution. The forward process generates a sequence of latent variables $\boldsymbol{z}_1, \boldsymbol{z}_2, \ldots, \boldsymbol{z}_T$ by successively introducing Gaussian noise, with distributions given by
\begin{equation}
q(\boldsymbol{z}_t|\boldsymbol{z}_{t-1}) = \mathcal{N}(\boldsymbol{z}_t; \sqrt{\alpha_t}\boldsymbol{z}_{t-1},
 (1-\alpha_t)\boldsymbol{I}) \, ,
\end{equation}
where $\alpha_t \in (0, 1)$ gives the variance schedule~\cite{Sohl-Dickstein_Weiss_Maheswaranathan_Ganguli_2015}. The marginal distribution of $\boldsymbol{z}_t$ given $\boldsymbol{z}_0$ is
\begin{equation}
q(\boldsymbol{z}_t|\boldsymbol{z}_0) = \mathcal{N}(\boldsymbol{z}_t; \sqrt{\bar{\alpha}_t}\boldsymbol{z}_0, (1-\bar{\alpha}_t)\boldsymbol{I})\, ,
\end{equation}
with $\bar{\alpha}_t = \prod_{s=1}^t \alpha_s$.  We will also write $\boldsymbol{z}_t = \sqrt{\bar{\alpha}_t} \boldsymbol{z}_0 + \sqrt{1-\bar{\alpha}_t} \boldsymbol{\epsilon}$, where $\boldsymbol{\epsilon} \sim \mathcal{N}(0, \boldsymbol{I})$\, .

\subsection{Reverse Diffusion Process}

The reverse diffusion process seeks to reconstruct the latent variable by reversing the forward process. This process is parameterised by a learnable set of parameters~$\theta$, and is conditioned on some context~$c$.  We take it to be
\begin{equation}
p_\theta(\boldsymbol{z}_{t-1}|\boldsymbol{z}_t, c) = \mathcal{N}(\boldsymbol{z}_{t-1}; \boldsymbol{\mu}_\theta(\boldsymbol{z}_t, t, c), (1-\alpha_t)\boldsymbol{I}) \, ,
\end{equation}
where the mean $\boldsymbol{\mu}_\theta(\boldsymbol{z}_t, t, c)$ is given by
\begin{equation}
\boldsymbol{\mu}_\theta(\boldsymbol{z}_t, t, c) = \frac{1}{\sqrt{\alpha_t}} \left(\boldsymbol{z}_t - \frac{1-\alpha_t}{\sqrt{1-\bar{\alpha}_t}}\boldsymbol{\epsilon}_\theta(\boldsymbol{z}_t, t, c)\right) \, ,
\end{equation}
and where $\boldsymbol{\epsilon}_\theta(\boldsymbol{z}_t, t, c)$ is learned to approximate~$\boldsymbol{\epsilon}$~\cite{Ho_Jain_Abbeel_2020}.

The training objective for diffusion models is based on the variational lower bound (VLB) of the negative log-likelihood of the data. However, it was shown in~\cite{Ho_Jain_Abbeel_2020} that a simpler objective could be used, which measures the difference between the true noise added and that predicted by the model
\begin{equation}
    \hat c = \argmin_c \mathcal{L}_{\text{diffusion}}(c) \, ,
\end{equation}
where $\boldsymbol{\epsilon} \sim \mathcal{N}(0, \boldsymbol{I})$, $t \sim \mathcal{U}[1,T]$ and $w_t = \min\{\frac{5}{\text{SNR}(t)}, 1\}$ is a time-dependent weighting function that balances the importance of different timesteps during training~\cite{Hang_2024, Ho_Jain_Abbeel_2020}.

\section{Training and Inference Parameters and Procedure}
\label{train_inf_setup}

We applied various data augmentations, including random diagonal flips, random rotation (with angles uniformly sampled between 0 and 359 degrees), colour jitter (brightness = 0.25, contrast = 0.25, saturation = 0.25, hue = 0.125), and Mixup~\cite{mixup} ($\alpha = 0.3$) applied to the conditioning instead of the target and RandAugment with default parameters~\cite{cubuk2020randaugment}. We used an AdamW optimiser ($\beta_1 = 0.9$, $\beta_2 = 0.999$, $\epsilon = 10^{-8}$, weight decay 0.01), mixed precision training (fp16), and an exponential moving average of 0.9999. All images were resized to $360 \times 360$ pixels.

During inference, we applied the same data augmentations used in training, excluding RandAugment and Mixup. Leveraging the fact that white blood cells are typically centred in our images, and to mitigate the effects of augmentations that corrupt the outer parts of the image, we calculated the inference error only for pixels within a radius of 20 from the centre of the image in our latent space, which has dimensions of $45 \times 45 \times 4$.

For comparisons with EfficientNetV2-M~\cite{tan2021efficientnetv2} and ViT-B/16~\cite{dosovitskiy2021image}, pre-trained on ImageNet~\cite{ImageNet}, we maintained consistent data augmentation techniques and optimiser settings. However, to accommodate the ViT-B/16 model's input size requirements of $384 \times 384$ pixels, we first resized the images to match our diffusion model's input size and then further resized them to $384 \times 384$ pixels. For these comparison models, we used a batch size of 16, a learning rate of $10^{-4}$, and trained for 50 epochs. The model checkpoint with the highest validation accuracy was selected for testing.

\section{CytoData Labelling Procedure}
\label{cytodata_labelling}

CytoData's labelling process involved six haematology experts with \{34, 25, 15, 12, 6, 5\} years of experience in blood microscopy, respectively. Each expert labelled 1,000 images, including 200 common images for inter-rater reliability assessment. Experts assigned one of ten classes (basophil, blast, lymphocyte, monocyte, neutrophil, erythroblast, eosinophil, immature granulocyte, platelet, or artefact) and indicated their confidence level (High, Moderate, Low, or No Confidence) for each prediction. For images labelled by multiple experts, we used a majority vote to determine the final label. In case of a tie, the label provided by the expert with the longest experience in blood microscopy was used.

The uncertainty label allows for in-depth comparisons between model and human expert performance, particularly in how certainty relates to accuracy. Researchers can investigate the patterns of confidence exhibited by models and human experts across different cases, and identify specific scenarios where models might demonstrate different certainty-accuracy relationships compared to human experts. Such analyses offer insights into the strengths and limitations of both AI models and human experts in haematological cell image classification. For a more detailed analysis of how labellers' confidence levels correlate with each other see Supplementary Figure~\ref{fig:confusion_matrix_labeller_confidence}. Table~\ref{tab:datasets} summarises the datasets used in our study, highlighting the unique attributes of CytoData.

To optimise the use of our expert labellers' time whilst also attempting to identify potential labelling errors, we employed a strategic review process. We trained our model using a 3-fold cross-validation approach, ensuring all images except the 200 common ones were used in the test set once. We then applied each model to its respective test set, effectively covering the entire dataset (excluding the 200 common images) with predictions from models that had not seen those images during training. Images for which the model's prediction did not agree with the original label were selected for review by two expert haematologists who had not previously labelled these specific images. This review process resulted in 816 images being evaluated by the two additional experts, of which 407 images had their class labels changed.

To implement this strategy, we continued the pretraining of a Stable Diffusion 1.5 model~\cite{Rombach_Blattmann_Lorenz_Esser_Ommer_2022} using a combination of ImageNet~\cite{ImageNet} and our unlabelled blood image dataset, with equal selection probability. The inclusion of ImageNet alongside our unlabelled blood images was crucial to prevent the model from unlearning its conditioning capabilities. Due to the large number of classes in ImageNet, which precluded the use of one-hot encoding, we employed binary encoded vectors for each class.

For this initial phase, we employed a batch size of 64 and trained on an A100-80GB GPU for 76,000 steps.
In the second phase, we started our training from this initial model to train our 3-fold cross-validation models for 45,000 steps each.

\section{Authenticity Test Setup} \label{turing_test_setup}

We fine-tuned the model on the combined PBC and Bodzas datasets, with all images resized to $512 \times 512$ pixels. From this combined dataset, we randomly selected 1,400 images for validation, with the remaining images allocated to the training set. The model was trained for 44,000 steps, employing a limited augmentation strategy that included only random vertical and horizontal flips.

Following the training phase, we generated a set of 144 synthetic images spanning all the classes using 50 inference steps and a guidance scale of 5, where the guidance scale controls the trade-off between image quality and adherence to the class condition~\cite{ho2022classifier}. Each image was conditioned on one of nine blood cell types: basophil, blast, lymphocyte, monocyte, neutrophil, erythroblast, eosinophil, immature granulocyte, and platelet. For the Bodzas dataset, we merged the neutrophil segment and neutrophil band classes into a single neutrophil class and created a blast class, which comprised both lymphoblasts and myeloblasts. To create a balanced dataset for the authenticity test, we randomly selected an equivalent set of 144 real images from our validation dataset, resulting in a total pool of 288 images.

\section{Psychometric Analysis Details} \label{psycho_analysis}

The ``threshold'' of the function ($m$)---an arbitrarily chosen point---and its slope or width ($w$) are ordinarily measures of performance relative to a ground truth signal strength. Since there is no ground truth here, the goodness of fit of the function to the data, and the separability of different observers are our measures of the quality of the uncertainty. The psychometric function is then defined as
\begin{equation}
\psi(x;m,w,\gamma,\lambda) = \lambda + (1 - \lambda - \gamma)S(x;m,w)\, ,
\end{equation}
where $S$ is a sigmoid, here chosen as the logistic, scaled in the interval [0,1]. We use Bayesian inference, implemented in psignifit (\url{https://github.com/wichmann-lab/psignifit} under MATLAB 2022b), to obtain parameter estimates, with priors informed by the psychophysical literature~\cite{schutt2016painfree}. For the threshold, we use a uniform prior over the range of the data with a cosine fall off to 0 over half the range of the data on either side of the maximum and the minimum, expressing a belief that the threshold is located with equal probability within the sampled range and may be up to 50\% of the spread of the data outside that range with decreasing probability. For the width, we use a uniform prior between two times the minimal distance of two tested stimulus levels and the range of the stimulus levels with cosine fall offs to 0 at the minimal difference of two stimulus levels and at 3 times the range of the tested stimulus.

Functions were fitted to CytoDiffusion performance with its uncertainty, with data from the entire test set; to individual expert performance with consensus expert uncertainty, with data from 200 cases for which consensus uncertainty was available; and to individual expert performance with model uncertainty, again with data from 200 cases for which consensus uncertainty was available. To provide sufficient support for each evaluated performance level, uncertainties were discretised into 25 bins for the CytoDiffusion function (where the full test set was available), and 12 bins for the expert functions (where 200 cases were available).

\section{Anomaly Detection Details} \label{anomaly_details}

We fine-tuned two diffusion models, one for each dataset, training each for 140,000 steps on an NVIDIA RTX A5000 GPU. For comparison, we also fine-tuned a Vision Transformer (ViT-B/16) model~\cite{dosovitskiy2021image} for each dataset.

For each image, we calculate the error for each class and normalise these errors by dividing by the mean error across all classes for that image. The confidence measure is then derived by computing the mean difference between the smallest error and all other errors, and subsequently subtracting the mean difference amongst all other errors. A larger score indicates higher confidence, as it suggests a greater distinction between the best-fitting class and the others.
For the ViT, we used the logits (before the softmax) instead of the errors.

\section{Ablation Studies on Weighting Functions and Pruning Hyperparameters} \label{ablation_weighting}

We conducted extensive ablation studies on different weighting functions and pruning hyperparameters to optimise our model's performance. Our investigations began with the `Heuristic' and `Learned' weighting functions described in \cite{Clark_Jaini_2023}. The `Heuristic' function takes the form $\exp(-kt)$, where $t$ is between 0 and 1, and $k$ was initially set to 6 or 7. The `Learned' function involved dividing the time interval into 20 buckets and fitting a logistic regression function to a small portion of the data to learn the weights.

Our findings revealed that whilst the median error scaled exponentially with time (with $k=5.5$ in our case, which performed better than 6 or 7), the standard deviation of the error did not follow this pattern. Instead, it formed a curve similar to a negative quadratic function. Given that our primary concern is relative differences rather than absolute values, we focused on this aspect. Consequently, we fitted a fourth-degree polynomial to the median of the standard deviation of the squared errors using validation data from the efficiency experiment with 20 images per class and used the reciprocal of this as our weighting function.
The coefficients of our polynomial function are
\begin{equation}
f(t) = 0.927 + 9.837t - 4.684t^2 + 6.019t^3
 - 9.340t^4\,
 ,
\end{equation}
and this weighting function was used consistently for all experiments presented in this paper.
Empirically, this approach outperformed even the fitted `Learned' logistic regression when using 240 images to train the logistic regression. Notably, our method generalised well, performing effectively on other datasets for which it was not specifically fitted.

We also explored alternative weighting functions. One approach involved calculating the error for all conditions $c$ for one $\boldsymbol{\epsilon}$ and $t$, normalising it to mean 0 and standard deviation 1, then drawing new $\boldsymbol{\epsilon}$ and $t$, normalising to $\mathcal{N}(0,1)$ etc. and taking the mean of all normalised errors. Another method ranked the errors, assigning a value of 1 to the lowest scoring error, 2 to the next lowest, and so on, before averaging across all trials. We also tested uniform weighting and the SNR weighting used for training: $\min\{\frac{5}{\text{SNR}(t)}, 1\}$.

To evaluate these weighting functions, we conducted ablation studies where we trained a model as described in the ``\nameref{Efficiency}'' section, with 20 images per class. During inference, we used \{1, 5, 10, 20, 50, 100, 250, 500, 1000\} trials per class without any pruning for each weighting function. Table~\ref{tab:accuracy_results} presents our results, showing the accuracy after 1000 trials and the mean accuracy across all trials. Our custom weighting function performed best, with the logistic regression and the normalised $\mathcal{N}(0,1)$ approaches following closely behind.

\begin{table}
\centering
\caption{Accuracy results for different weighting strategies.}
\label{tab:accuracy_results}
\begin{tabular}{
  l
  S[table-format=1.4]
  S[table-format=1.4]
}
\toprule
{Weighting Strategy} & {Accuracy after 1000 trials} & {Mean Accuracy} \\
\midrule
Custom polynomial    & 0.9504 & 0.8805 \\
Logistic regression  & 0.9458 & 0.8761 \\
Normalised $\mathcal{N}(0,1)$ & 0.9504 & 0.8743 \\
Uniform              & 0.9371 & 0.8704 \\
SNR                  & 0.9371 & 0.8704 \\
Ranking              & 0.9467 & 0.8689 \\
$\exp(-5.5t)$        & 0.9287 & 0.8642 \\
\bottomrule
\end{tabular}
\end{table}

\begin{figure}[ht]
    \centering
    \includegraphics[width=1.0\linewidth]{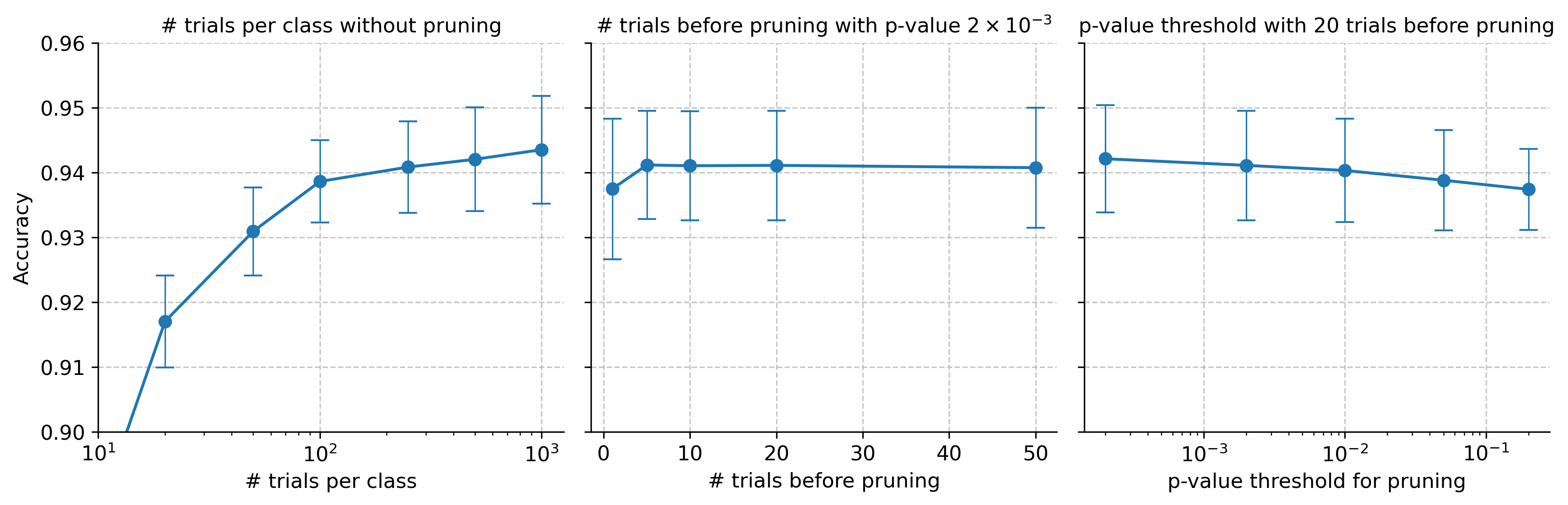}
    \caption{Ablation studies on pruning hyperparameters. Left: Performance vs number of trials per class (no pruning). Middle: Impact of delaying pruning initiation. Right: Effect of p-value threshold on pruning.}
    \label{fig:ablation_pruning}
\end{figure}

We further investigated the impact of the p-value threshold for our pruning and the number of trials to conduct before initiating the Student's $t$-test to determine if any class should be pruned. Figure~\ref{fig:ablation_pruning} illustrates these results, showing the mean and standard deviation across all weighting functions combined.

The left panel of Figure~\ref{fig:ablation_pruning} demonstrates that performance continues to improve with the number of trials per class, even beyond 1000 trials (no pruning was used here). The middle panel shows the effect of delaying the start of pruning until a certain number of trials have been conducted (to avoid prematurely pruning the correct class). Here, we used a fixed p-value threshold of $2\times10^{-3}$. We observe that as long as we use $\geq5$ steps, the performance is not significantly affected.

The right panel illustrates the impact of the p-value threshold when pruning begins after 20 steps. As expected, performance improves with lower thresholds, approaching the theoretical maximum (the highest value in the left panel) when a value of $2\times10^{-4}$ is used.

\section{Generated Images}
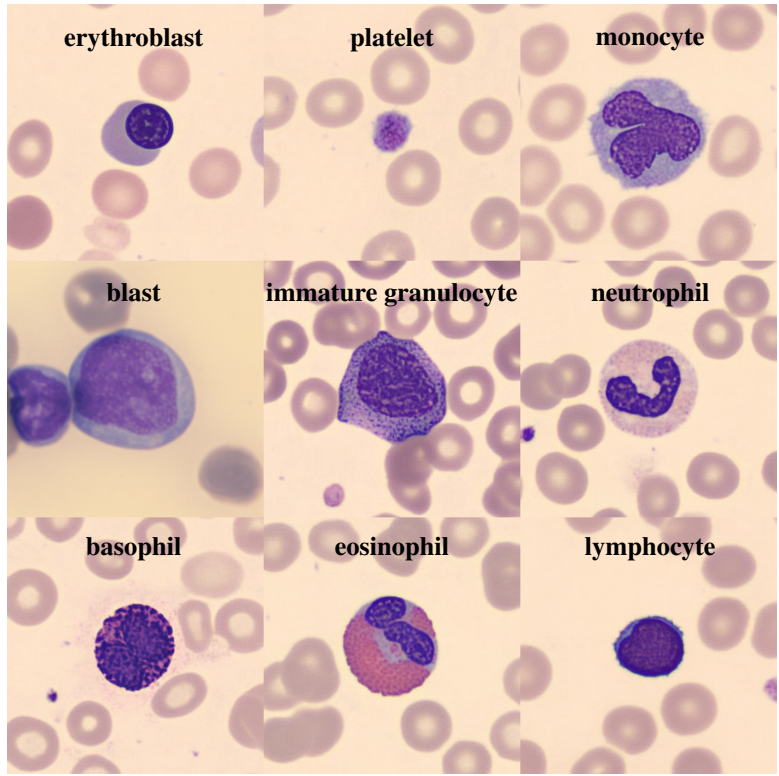
\begin{figure}[ht]
\centering
\begin{tikzpicture}
\node[anchor=south west,inner sep=0] (image) at (0,0) {\includegraphics[width=0.62\textwidth]{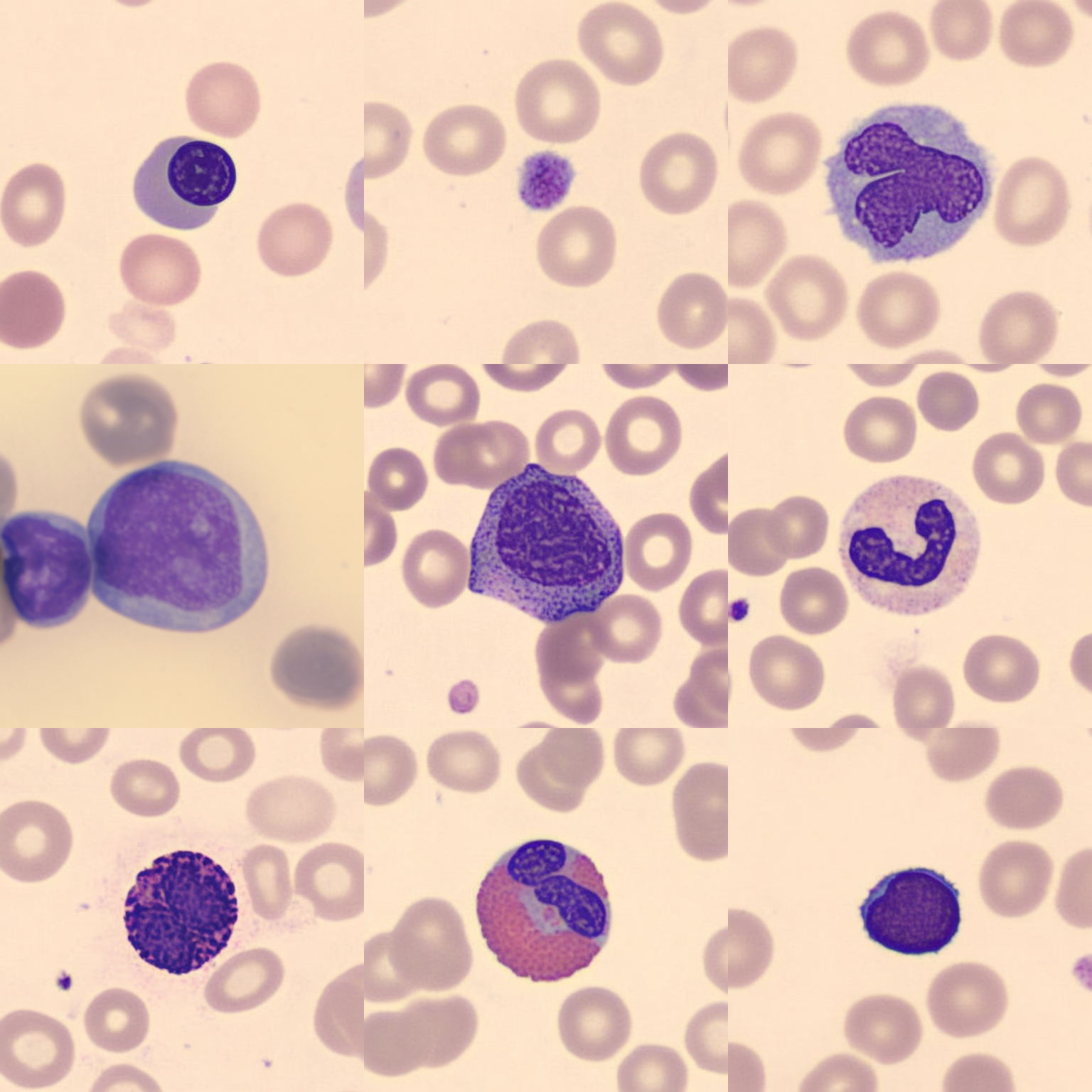}};
\begin{scope}[x={(image.south east)},y={(image.north west)}]
    \node[anchor=north] at (0.165,0.98) {\textbf{erythroblast}};
    \node[anchor=north] at (0.5,0.98) {\textbf{platelet}};
    \node[anchor=north] at (0.835,0.98) {\textbf{monocyte}};
    \node[anchor=north] at (0.165,0.65) {\textbf{blast}};
    \node[anchor=north] at (0.5,0.65) {\textbf{immature granulocyte}};
    \node[anchor=north] at (0.835,0.65) {\textbf{neutrophil}};
    \node[anchor=north] at (0.165,0.32) {\textbf{basophil}};
    \node[anchor=north] at (0.5,0.32) {\textbf{eosinophil}};
    \node[anchor=north] at (0.835,0.32) {\textbf{lymphocyte}};
\end{scope}
\end{tikzpicture}
\caption{Synthetic blood cell images generated by our fine-tuned diffusion model. The figure displays one synthetic image for each of the nine blood cell classes. These images demonstrate the model's ability to capture distinct morphological features of each cell type.}
\label{fig:synthetic-blood-cells}
\end{figure}

The images presented in Figure~\ref{fig:synthetic-blood-cells} showcase the capability of our fine-tuned diffusion model to generate highly realistic synthetic blood cell images. Each image in the grid represents one blood cell class. These synthetic images were part of the dataset used in the authenticity test described in the main text, where expert haematologists were unable to reliably distinguish them from real blood cell images.

\section{Labeller Confidence Analysis}

\begin{figure}[ht]
    \centering
    \includegraphics[width=0.55\linewidth]{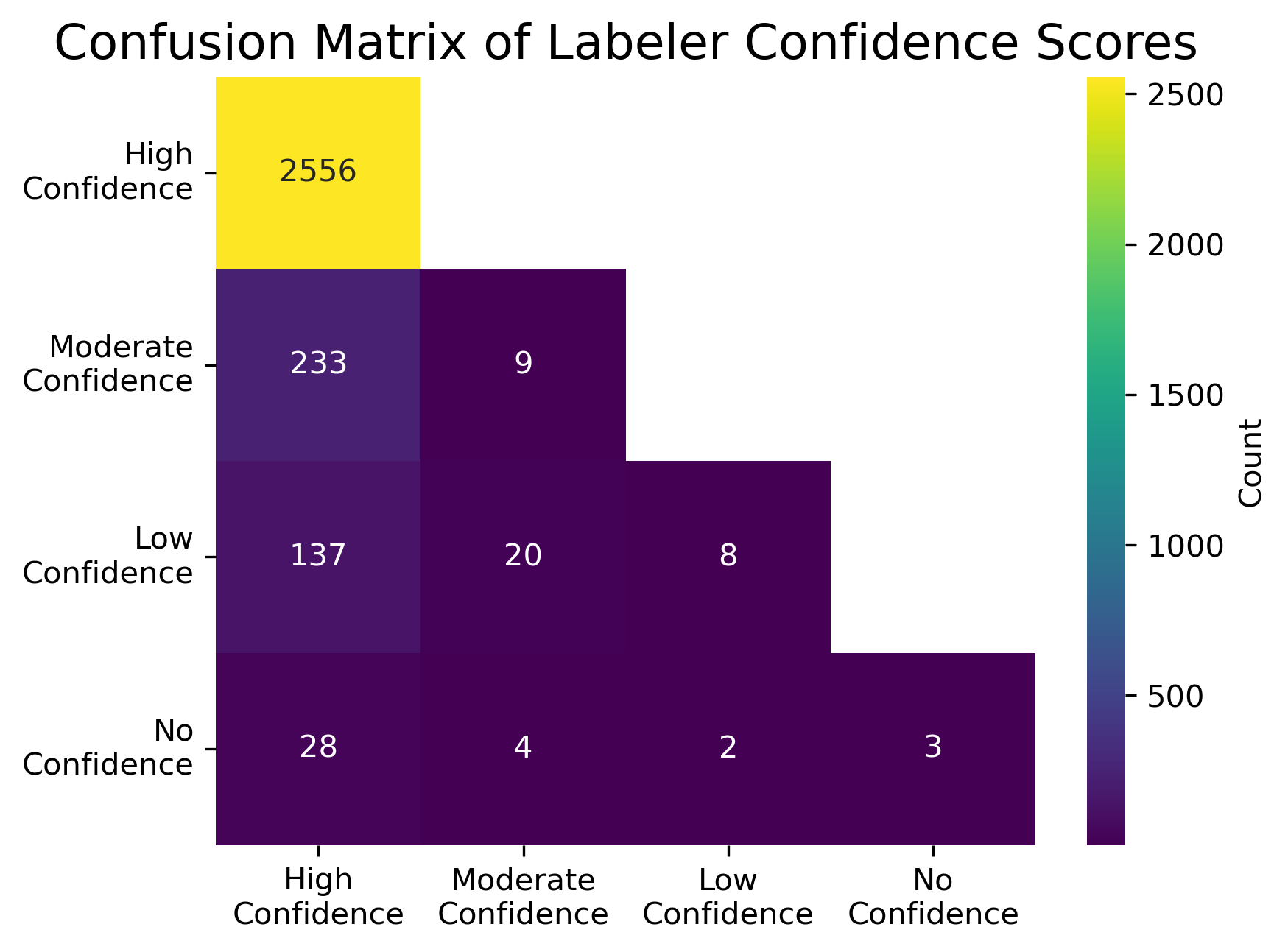}
    \caption{Confusion matrix of labeller confidence scores. The matrix shows the frequency of agreement and disagreement between different labellers' confidence levels when assessing the same 200 images. Each cell represents the count of pairwise comparisons (out of 3,000 total comparisons) where the confidence levels of two labellers aligned as indicated. The diagonal represents agreement, whilst off-diagonal elements indicate discrepancies in confidence assessments between labellers.}
    \label{fig:confusion_matrix_labeller_confidence}
\end{figure}

To provide insight into the consistency of labeller confidence across different experts, we analysed the agreement between labellers when assessing the same images. Figure~\ref{fig:confusion_matrix_labeller_confidence} presents a confusion matrix of labeller confidence scores based on a subset of our dataset.

For this analysis, we used a set of 200 images that were independently labelled by all six expert haematologists in our study. We then evaluated their confidence levels pairwise, resulting in 30 comparisons per image.

The matrix reveals a strong tendency for labellers to agree on high-confidence assessments, as evidenced by the large number (2,556) in the top-left cell. However, there is also notable disagreement in some cases, particularly between high and moderate confidence levels.

Interestingly, instances of `No Confidence' are relatively rare and often do not align with other labellers' assessments. This could indicate that factors leading to a complete lack of confidence are highly subjective or dependent on individual expertise.

These findings underscore the complexity of expert assessment in haematological image analysis and highlight the potential value of incorporating confidence measures in models to go beyond expert human behaviour.

\section{High-Level Explanation of Model Prediction} \label{High-level}

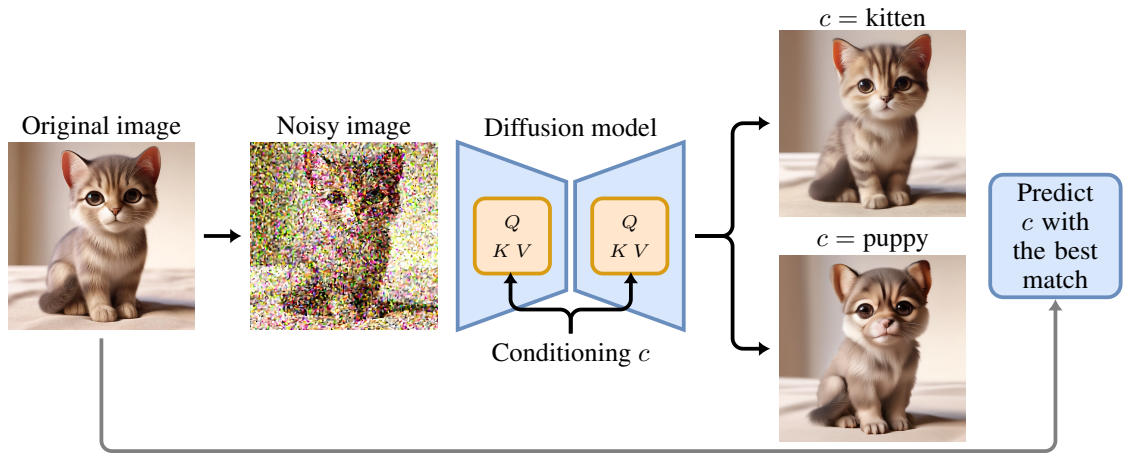
\begin{figure}[htp]
    \centering
    \begin{adjustbox}{center}
    \begin{tikzpicture}
    \definecolor{light_blue}{rgb}{0.847, 0.906, 0.984}
    \definecolor{blue}{rgb}{0.412, 0.553, 0.741}
    \definecolor{boxfill}{RGB}{255, 230, 205}
    \definecolor{boxedge}{RGB}{216, 155, 18}

    \node (original) {\includegraphics[width=0.15\textwidth]{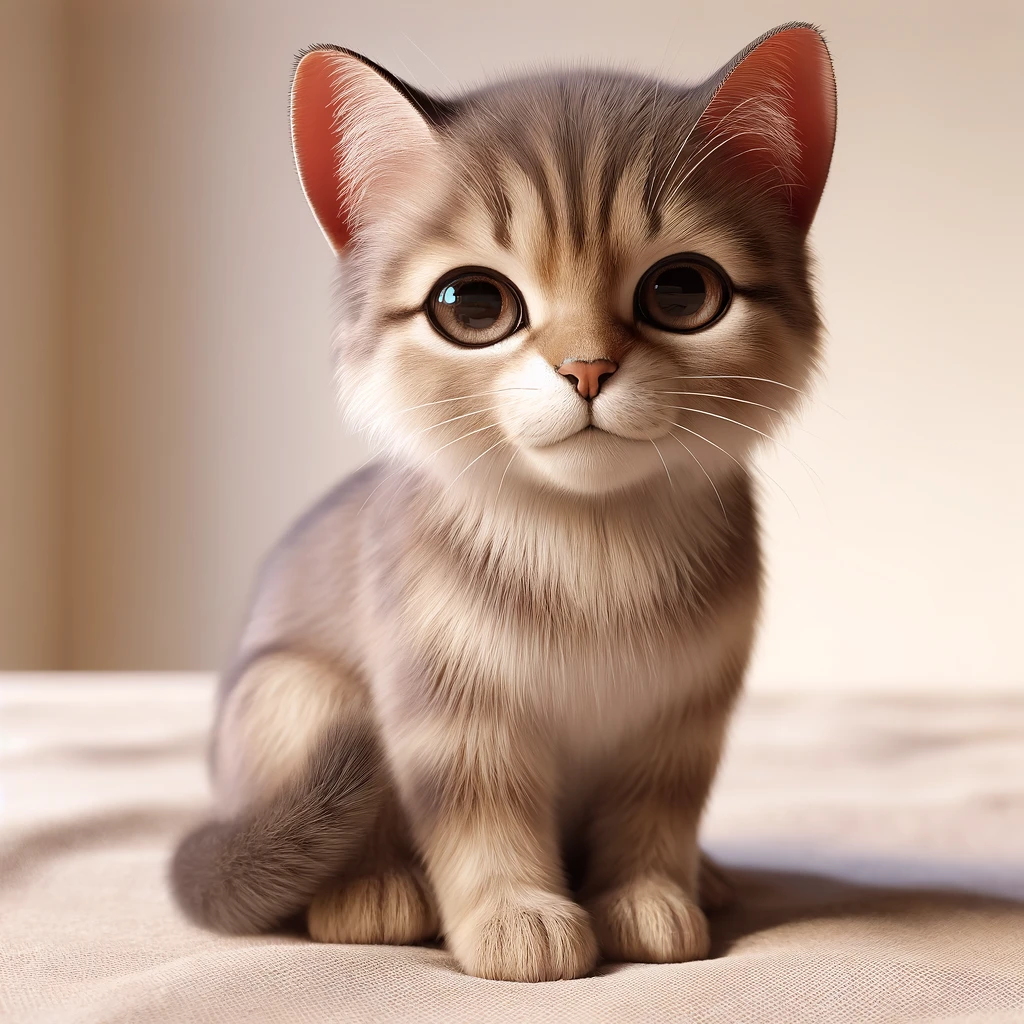}};

    \node[above=of original, yshift=-1.2cm] (label_original) {Original image};

    \node[right=of original, xshift=-0.50cm] (combined) {\includegraphics[width=0.15\textwidth]{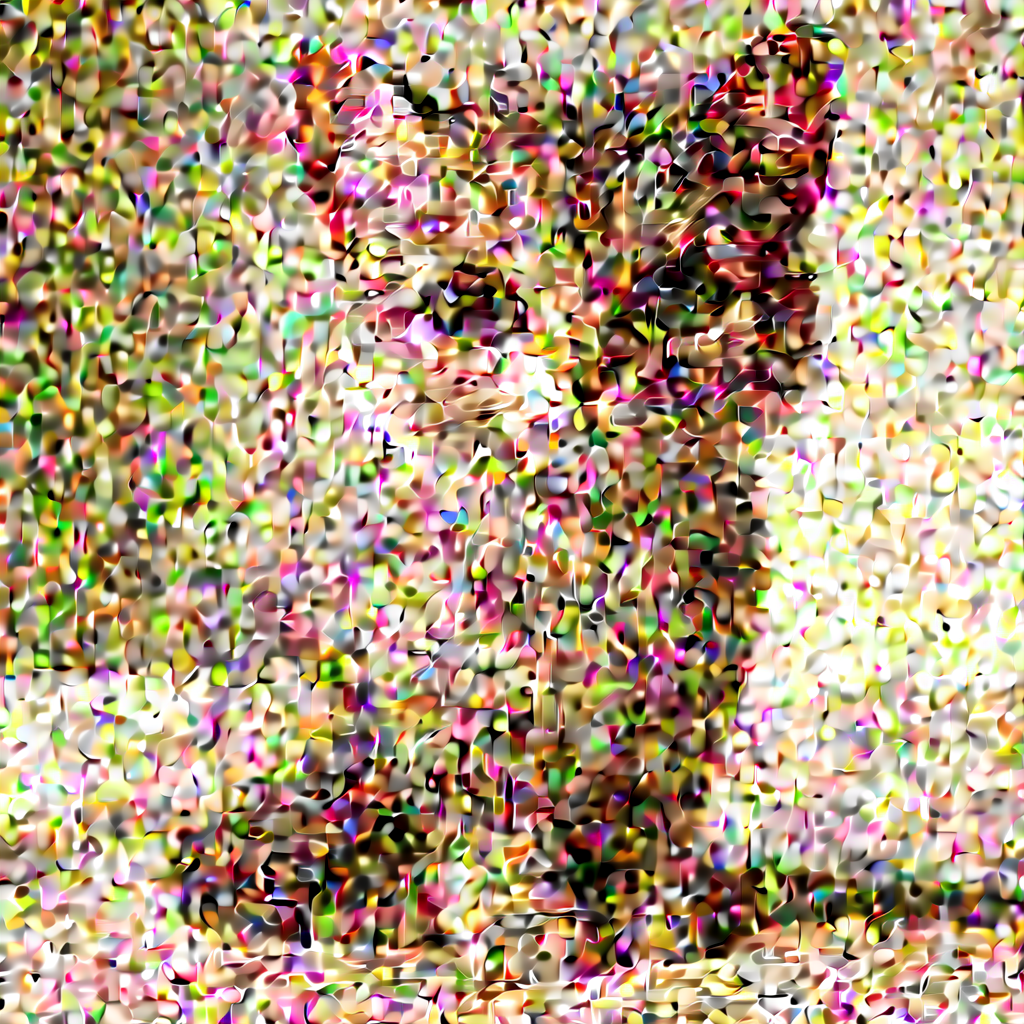}};

    \node[above=of combined, yshift=-1.2cm] (xt) {Noisy image};

    \node[right=of combined, xshift=-1.0cm, yshift=0.0cm] (stable_diffusion) {
    \begin{tikzpicture}

        \node[trapezium, trapezium angle=70, minimum width=2.5cm, minimum height=0.47cm, draw=blue, fill=light_blue, line width=0.5mm, shape border rotate=270] (encoder_SD) {};
        \node[trapezium, trapezium angle=70, minimum width=2.5cm, minimum height=0.47cm, draw=blue, fill=light_blue, line width=0.5mm, shape border rotate=90, right=0.067cm of encoder_SD] (decoder_SD) {};

        \node[rectangle, rounded corners, draw=boxedge, fill=boxfill, minimum width=1.0cm, minimum height=1.0cm, align=center, line width=0.5mm, at=(encoder_SD.center), xshift=-0.52cm] (QKV_e) {\scriptsize $Q$\\\scriptsize$K$ \scriptsize$V$};
        \node[rectangle, rounded corners, draw=boxedge, fill=boxfill, minimum width=1.0cm, minimum height=1.0cm, align=center, line width=0.5mm, at=(decoder_SD.center), xshift=-0.52cm] (QKV_d) {\scriptsize $Q$\\\scriptsize$K$ \scriptsize$V$};

    \end{tikzpicture}
    };
    \node[below=of stable_diffusion, yshift=1.1cm, xshift=0.0cm] (class_conditioning) {Conditioning $c$};

    \draw[-{Latex[length=1.5mm, width=2mm]}, line width=0.5mm, rounded corners=4pt]
        (class_conditioning)
        -- ++(0,0.6)
        -- ++(-0.8,0)
        -- ++(0,0.5);
    \draw[-{Latex[length=1.5mm, width=2mm]}, line width=0.5mm, rounded corners=4pt]
        (class_conditioning)
        -- ++(0,0.6)
        -- ++(0.8,0)
        -- ++(0,0.5);

    \node[above=of stable_diffusion, yshift=-1.2cm] (SD_label) {Diffusion model};

    \node[right=of stable_diffusion, xshift=0.0cm, yshift=1.5cm] (kitten_recon) {\includegraphics[width=0.15\textwidth]{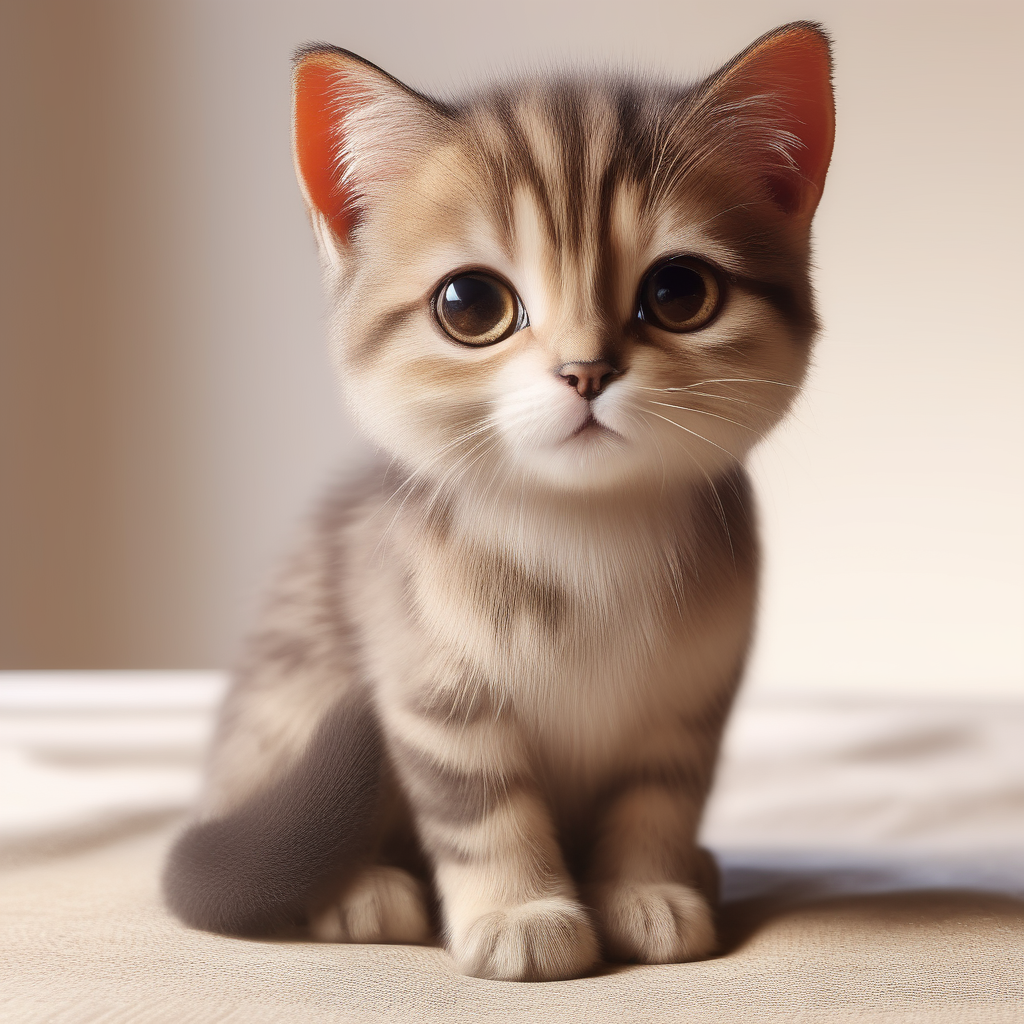}};

    \node[above=of kitten_recon, yshift=-1.2cm] (label_kitten_recon) {$c=\text{kitten}$};

    \node[right=of stable_diffusion, xshift=0.0cm, yshift=-1.5cm] (kitten_puppy) {\includegraphics[width=0.15\textwidth]{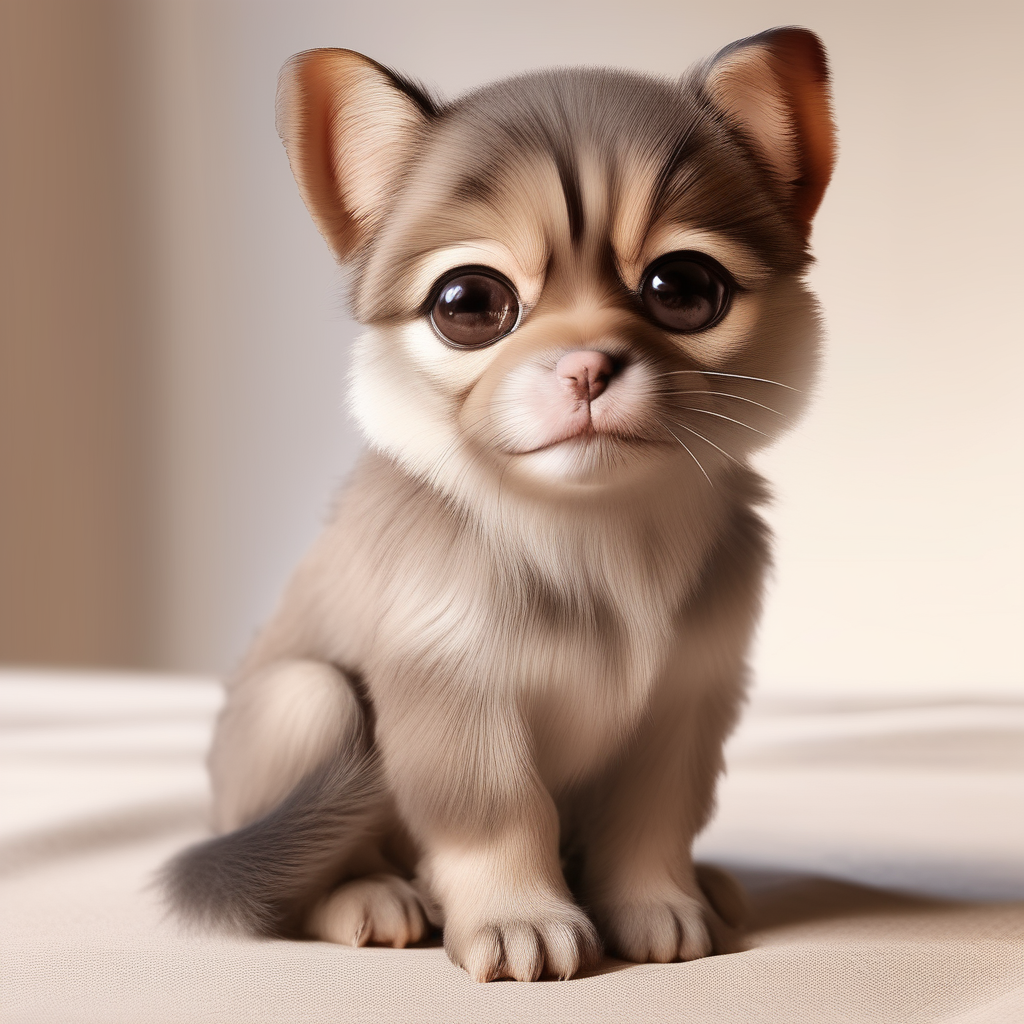}};

    \node[above=of kitten_puppy, yshift=-1.2cm] (label_kitten_recon) {$c=\text{puppy}$};

    \node[right=of stable_diffusion, xshift=2.9cm, text width=1.5cm, line width=0.5mm, align=center, fill=light_blue, draw=blue, rounded corners=5pt] (note) {Predict $c$ with the best match};

    \draw[-{Latex[length=1.5mm, width=2mm]}, line width=0.5mm, rounded corners=4pt] (original) -- (combined);

        \draw[-{Latex[length=1.5mm, width=2mm]}, line width=0.5mm, rounded corners=4pt]
    (stable_diffusion.east)
    -- ++(0.5,0)

    -- ++(0,1.5)
    -- (kitten_recon.west);

    \draw[-{Latex[length=1.5mm, width=2mm]}, line width=0.5mm, rounded corners=4pt]
        (stable_diffusion.east)
        -- ++(0.5,0)
        -- ++(0,-1.5)
        -- (kitten_puppy.west);

    \draw[-{Latex[length=1.5mm, width=2mm]}, line width=0.5mm, rounded corners=4pt, draw=gray] (original.south) ++(0,-0.0) -- ++(0.0,-1.5) -- ++(10,-0.0)  -| (note.south);

    \end{tikzpicture}
    \end{adjustbox}
    \caption{High-level overview of how the model predicts a class.}
    \label{fig:kitten}
\end{figure}

For an intuitive explanation of how our model makes its predictions, we refer to Figure~\ref{fig:kitten}. This figure presents a toy example demonstrating how the model determines whether an original image depicts a kitten or a puppy.

The process begins by adding noise to the original image. This noisy image is then fed through the model twice: once with the condition that it is a kitten (that is, the model is told that it is an image of a kitten), and once with the condition that it is a puppy. For each condition, the model predicts how to transform the noisy image into an image of either a kitten or a puppy, respectively.

The model then compares the original image with the created images for the different classes. The class that produces the image most closely matching the original is selected as the model's prediction.

\end{document}